%% file: main.tex
\documentclass[10pt,twocolumn,letterpaper]{article}

\usepackage[pagenumbers]{cvpr} 

\usepackage{graphicx}
\usepackage{float} 
\usepackage{caption}
\usepackage{amsmath}
\usepackage{amssymb}
\usepackage{booktabs}
\usepackage{multirow}
\usepackage{enumitem}
\usepackage{xcolor, color, colortbl}
\usepackage{pbox} 
\usepackage{tabularx} 
\usepackage{pythonhighlight} 
\lstnewenvironment{PythonA}[1][]{\lstset{style=mypython, #1}}{}

\definecolor{cvprblue}{rgb}{0.21,0.49,0.74}

\newcolumntype{a}{>{\columncolor{Gray}}c}
\newcolumntype{b}{>{\columncolor{white}}c}
\usepackage[pagebackref,breaklinks,colorlinks,citecolor=cvprblue]{hyperref}
\input{preamble}

\usepackage[capitalize]{cleveref}
\crefname{section}{Sec.}{Secs.}
\Crefname{section}{Section}{Sections}
\Crefname{table}{Table}{Tables}
\crefname{table}{Tab.}{Tabs.}

\newcommand{\website}{https://snap-research.github.io/AToM}

\def\cvprPaperID{2855} 
\def\confName{CVPR}
\def\confYear{2024}

\begin{document}


\title{%
  \includegraphics[height=1em]{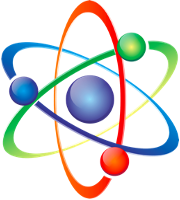}
  AToM: Amortized Text-to-Mesh using 2D Diffusion
}

\author{
Guocheng Qian$^{1,2}$
\quad Junli Cao$^{1}$ 
\quad Aliaksandr Siarohin$^{1}$
\quad Yash Kant$^{1,3}$ 
\quad Chaoyang Wang$^{1}$ \\
Michael Vasilkovsky$^{1}$ 
\quad Hsin-Ying Lee$^{1}$ 
\quad Yuwei Fang$^{1}$
\quad Ivan Skorokhodov$^{1}$ \quad Peiye Zhuang$^{1}$ \\
Igor Gilitschenski$^{3}$ \quad Jian Ren$^{1}$ \quad Bernard Ghanem$^{2}$ \quad Kfir Aberman$^{1}$ \quad Sergey Tulyakov$^{1}$ \\
$^{1}$Snap Research \quad $^{2}$KAUST \quad $^{3}$University of Toronto
}


\makeatletter
\let\@oldmaketitle\@maketitle
\renewcommand{\@maketitle}{\@oldmaketitle
\myfigure\bigskip}
\makeatother
\newcommand\myfigure{%
  \makebox[0pt]{\hspace{17.5cm}\includegraphics[width=1.0\textwidth]{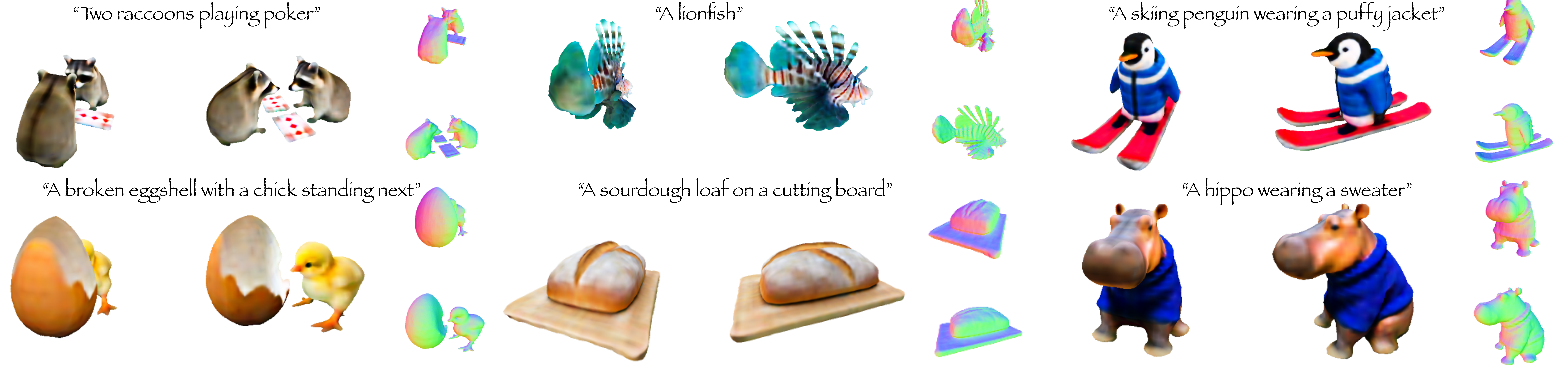}}
  \\
\refstepcounter{figure}\normalfont{Figure~\thefigure: Our Amortized Text-to-Mesh model (AToM), trained only on 2D diffusion prior, can generate textured meshes from texts in \textbf{less than 1 second}. See \url{\website} for immersive visualization. }
  \label{fig:teaser}
}

\maketitle

\input{sections/abstract}

\input{sections/introduction}
\input{sections/relatedwork}
\input{sections/methodology}

\input{sections/experiments}
\input{sections/ablation}

\input{sections/conclusion}
\input{sections/supplement}

\include{figures/gallery2}

{\small
\bibliographystyle{ieee_fullname}
\bibliography{main}
}

\end{document}

%% file: preamble.tex
\usepackage{bm}

\newcommand{\figlabel}{Fig.\xspace}

\newcommand{\tablabel}{Tab.\xspace}
\newcommand{\inlinesection}[1]{\noindent\textbf{#1.}}

\newcommand{\supp}{\textit{Appendix}\xspace}

\definecolor{Highlight}{HTML}{39b54a}  

\usepackage{amssymb}
\usepackage{pifont}
\definecolor{gr}{gray}{.95}

\definecolor{demphcolor}{gray}{0.5}
\newcommand{\demph}[1]{\textcolor{demphcolor}{#1}}

\usepackage{algorithm}
\usepackage{listings}
\usepackage{etoolbox}
\makeatletter
\AfterEndEnvironment{algorithm}{\let\@algcomment\relax}
\AtEndEnvironment{algorithm}{\kern2pt\hrule\relax\vskip3pt\@algcomment}
\let\@algcomment\relax
\newcommand\algcomment[1]{\def\@algcomment{\footnotesize#1}}
\renewcommand\fs@ruled{\def\@fs@cfont{\bfseries}\let\@fs@capt\floatc@ruled
  \def\@fs@pre{\hrule height.8pt depth0pt \kern2pt}%
  \def\@fs@post{}%
  \def\@fs@mid{\kern2pt\hrule\kern2pt}%
  \let\@fs@iftopcapt\iftrue}
\makeatother
\newcommand{\cmmnt}[1]{}

\usepackage[normalem]{ulem}

%% file: sections/abstract.tex
\begin{abstract}
We introduce Amortized Text-to-Mesh (AToM), a feed-forward text-to-mesh framework optimized across multiple text prompts simultaneously. In contrast to existing text-to-3D methods that often entail time-consuming per-prompt optimization and commonly output representations other than polygonal meshes, AToM directly generates high-quality textured meshes in less than 1 second with around $10\times$ reduction in the training cost, and generalizes to unseen prompts. Our key idea is a novel triplane-based text-to-mesh architecture with a two-stage amortized optimization strategy that ensures stable training and enables scalability. Through extensive experiments on various prompt benchmarks, AToM significantly outperforms state-of-the-art amortized approaches with over $4\times$ higher accuracy (in DF415 dataset) and produces more distinguishable and higher-quality 3D outputs. AToM demonstrates strong generalizability, offering finegrained 3D assets for unseen interpolated prompts without further optimization during inference, unlike per-prompt solutions. 
\end{abstract}

%% file: sections/introduction.tex
\section{Introduction}
\label{sec:intro}

Polygonal meshes constitute a widely used and efficient representation of 3D shapes. As we enter a revolutionary phase of Generative AI \cite{DiffusionModels, LDM}, the creation of 3D meshes has become increasingly intuitive, with controls transitioning from complex, handcrafted graphics handles \cite{pixel2mesh} to simple textual inputs \cite{Get3D}. 
Current mainstream text-to-mesh models \cite{Magic3D,Fantasia3D,TextMesh} can  generate impressive textured meshes through score distillation sampling \cite{DreamFusion} without 3D supervision. 
Despite the growing interest, these methods require a \textit{per-prompt optimization} that trains a standalone model for each prompt, which is time and computational consuming. 
More importantly, per-prompt solutions \textit{cannot generalize to unseen prompts}.

Recently, ATT3D \cite{ATT3D} presents amortized text-to-3D, which optimizes a text-to-3D system in many prompts simultaneously unlike per-prompt solutions. This amortized optimization not only significantly reduces training time but also allows generalizability due to the feature sharing across prompts. 
Unfortunately, ATT3D is limited predominantly to outputting 3D objects represented by Neural Radiance Fields (NeRF) \cite{NeRF}. 
An amortized text-to-mesh system is of more practical importance, but is under explored.
First, mesh is more widely used in most developments such as gaming and design. However, converting NeRFs to meshes is inaccurate and might require further optimization that is costly \cite{MarchingCubes}. 
Second, training text-to-mesh directly facilities a more efficient rasterizer, allowing higher-resolution renders that help recover details in geometry and texture compared to text-to-NeRF \cite{Magic3D,Fantasia3D}.

Extending ATT3D to amortized mesh generation presents challenges in unstable training that causes poor geometry. Our observations highlight two primary factors contributing to this \textbf{instability of ATT3D for mesh generation}: the \textbf{architecture} and the \textbf{optimization}.
\textit{First}, ATT3D adopted a HyperNetwork-based\cite{HyperNetworks} Instant-NGP \cite{InstantNGP} positional encoding for text-conditioned NeRF generation. This HyperNetwork introduces numerical instability and demands special layers such as spectral normalization \cite{SpectralNorm} to alleviate. 
The instability is more severe in large-scale datasets, leading to indistinguishable 3D content for different prompts, limiting generalizability of the system. See the two similar robots generated from two distinct prompts in $4^{th}$ row $1^{st}$ and $3^{rd}$ column in \figlabel \ref{fig:sota-df415}. 
\textit{Second}, the end-to-end optimization for text-to-mesh also triggers instability due to the topological limitations of differential mesh representation \cite{DMTet}, leading to suboptimal geometry.
Overall, naively extending ATT3D to mesh generation results in divergent optimization and the inability to generate any 3D object after training. 
Refer to the second column 1 in \figlabel\ref{fig:pullfig} for illustration.

\input{figures/pullfigure}

We thus introduce \textit{AToM}, the first \textit{a}mortized approach for direct \textit{T}ext-to-\textit{M}esh generation. 
To address architecture instability, AToM introduces a \textit{text-to-triplane} network in replacement of HyperNetwork for the positional encoding. 
Our text-to-triplane demonstrates greater resilience to parameter changes and generally yields higher-quality and significantly more distinguishable 3D content compared to the ATT3D's HyperNetwork counterpart.
We then propose to use triplane features as input to subsequent signed distance function (SDF), deformation, and color networks to generate geometry and texture for differentiable mesh \cite{DMTet}. 

Moreover, to stabilize optimization, we propose a two-stage amortized training in contrast to naive end-to-end optimization. 
Our first stage trains text-to-triplane, SDF, and color networks through low-resolution ($64\times64$) volumetric rendering. 
Volumetric rendering's consideration of multiple points per ray contributes to a stable optimization of the SDF network. In our second stage, these networks undergo refinement, and an additional deformation network is learned to manipulate the mesh vertices for finegrained details. Utilizing efficient mesh rasterization allows for $512\times512$ resolution renders in this phase. After training, AToM enables ultra-fast inference, generating textured meshes in under one second.
The main \textbf{contributions} of this work can be summarized as follows:
\begin{itemize}[leftmargin=10pt,nosep]
\item We propose AToM, the first amortized text-to-mesh model that is optimized across multiple text prompts without 3D supervision. AToM trains a triplane-based mesh generator, which contributes to stable optimization and generalizability to large-scale datasets.

\item We introduce a two-stage amortized optimization, where the first stage uses low-resolution volumetric rendering, and the second stage utilizes high-resolution mesh rasterization. Our two-stage amortized training significantly improves the quality of the textured mesh. 

\item AToM generates high-quality textured meshes in less than $1$ second from a text prompt and generalizes to unseen prompts with no further optimization.

\end{itemize}

%% file: figures/pullfigure.tex
\begin{figure}[t]
\centering
\includegraphics[page=1,width=0.5\textwidth, trim=0.1cm 15.4cm 21.7cm 0, clip]{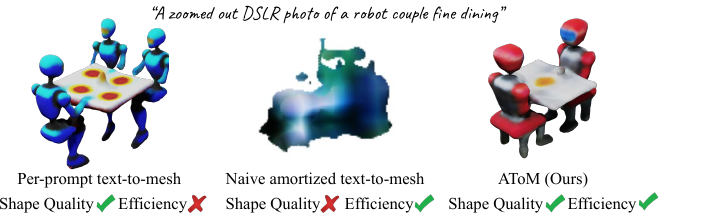}
\caption{
Per-prompt text-to-mesh \cite{TextMesh} generates high-quality results but demands expensive optimization. Naively extending ATT3D for mesh generation leads to divergent training and poor geometry. AToM introduces a triplane-based architecture with two-stage amortized optimization for enhanced stability. AToM efficiently generates textured meshes for various text prompts in under one second during inference.
}
\label{fig:pullfig}
\end{figure}

%% file: sections/relatedwork.tex
\section{Related Work}\label{sec:related}

\input{figures/pipeline}

\inlinesection{Feed-Forward 3D Generation}
The evolution of feed-forward 3D generation models has followed the success of 2D generation, drawing inspiration from generative adversarial networks~\cite{GAN,StyleGAN,pigan} and autoregressive networks~\cite{3dilg,autosdf} to diffusion models~\cite{Point-E,Shap-E,SDFusion,3dgen,SSDNerf,3dshape2vecset}. 
Various 3D representations have been studied, including point clouds~\cite{pcGAN_wu,pcGAN_li,pcGAN_shu}, volumes~\cite{infinicity,smith2017improved,xie2018learning,crossmodel3d}, and meshes~\cite{sketch2model,chen2019learning,pavllo2020convolutional,pixel2mesh}.
Despite their success, these methods are bounded by the availability of high-quality 3D data and thus most previous works merely applied to certain categories, such as cars and human faces \cite{Get3D,EG3D}. 
The concurrent work Instant3D \cite{Instant3D} shows the potential to train a generalizable 3D generative model in the recent large-scale 3D dataset \cite{Objaverse-XL}. We note that training in 3D dataset or through score distillation sampling (SDS) \cite{DreamFusion} are two orthogonal directions. The latter does not require any 3D data, which aligns with our interest. We elaborate the text-to-3D by SDS next.

\inlinesection{Per-prompt 3D Optimization} 
Recent studies have suggested leveraging pretrained text-to-image models for 3D generation without 3D data supervision ~\cite{DreamFields, DreamFusion, RealFusion, Magic123, scenetex}. Subsequent works introduced enhancements in directions such as multiview image gradient aggregation~\cite{SJC}, two-stage training optimization~\cite{Magic3D}, representation techniques~\cite{Fantasia3D}, increased diversity~\cite{ProlificDreamer}, and optimization techniques~\cite{HiFA}. Being able to generate high-quality 3D content, per-prompt optimization is receiving increasing interest. However, these methods are time-intensive, as each asset needs a separate optimization process and usually requires tedious parameter tuning. Per-prompt optimization also overfits to the single training prompt. Instead, we are interested in generalizable and efficient text-to-3D. 

\inlinesection{Amortized Optimization}
Unlike the time-consuming per-prompt optimization, ATT3D \cite{ATT3D} proposed to amortize \cite{amortizedop} the optimization across multiple prompts. This enables more efficient synthesis of 3D objects in seconds, facilitating interpolation between prompts and generalization to unseen prompts. However, ATT3D is limited to small-scale datasets, generating 3D content indistinguishable between prompts in larger-scale benchmark, \eg DF415 (415 prompts from DreamFusion \cite{DreamFusion}). Additionally, ATT3D solely produces NeRF that limits the quality.  A recent concurrent work HyperFields \cite{HyperFields} attempted to improve ATT3D with a stronger dynamic hypernetwork. In this work, we are more interested in amortized text-to-mesh, that generates textured meshes in under one second and can be applied to large-scale prompt datasets.

%% file: figures/pipeline.tex
\begin{figure*}[tbh]
\centering
\includegraphics[page=1,width=1.0\textwidth, trim= 0 9.6cm 10.5cm 0cm, clip]{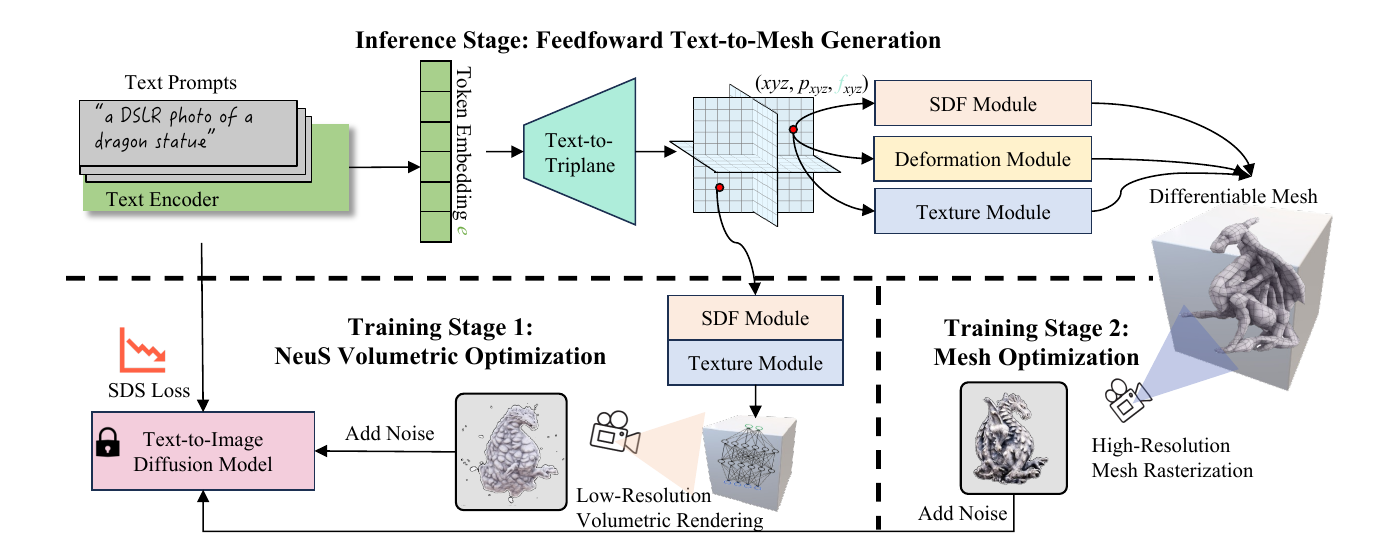}
\caption{\textbf{Inference and training of AToM.} \textbf{AToM inference (up)}: AToM generates textured meshes from given prompts in less than a second in inference. The text-to-mesh generator proposed in AToM consists of three components: a) a text encoder that tokenizes the input prompt, b) a text-to-triplane network that outputs a triplane representation from the text embedding, and c) a 3D network that generates SDF, vertex deformation, and color to form a differential mesh from positions and triplane features. 
\textbf{AToM Training (bottom)}: AToM utilizes a two-stage amortized optimization, where the first stage leverages stable volumetric optimization to train only the SDF and texture modules using low-resolution renders. The seconds stage uses mesh rasterization to optimize the whole network through high-resolution renders.
In both stages, AToM is trained simultaneously on many prompts through the guidance of a text-to-image diffusion prior without any 3D data supervision.
}
\label{fig:pipeline}
\end{figure*}

%% file: sections/methodology.tex
\section{Method}\label{sec:methodology}

\subsection{AToM Pipeline}\label{sec:pipeline}
\figlabel\ref{fig:pipeline} demonstrates the pipeline of the proposed Amortized Text-to-Mesh (AToM). 
Unlike mainstream per-prompt solutions \cite{DreamFusion, Magic3D, TextMesh, Fantasia3D} that train a standalone 3D model for a specific prompt, AToM trains a text-conditioned mesh generator, which can efficiently produce textured meshes from various text prompts in inference stage.
The network architecture of AToM consists of three components: (1) a text encoder, (2) a text-to-triplane network, and (3) a triplane-to-mesh generator.

\noindent\textbf{Text encoder} embeds the input text prompt. For simplicity, we use the same frozen pretrained T5 XXL \cite{T5} as  the text-to-image diffusion model DeepFloyd IF \cite{DeepFloyd}. 
The text embedding  $e \in \mathbb{R}^{L_e\times C_e}$ obtained is used as input to the following networks to generate the desired 3D content, where $L_e$ and $C_e$ represent the number of tokens (\eg 77) and the dimension of their embedding (\eg 4096), respectively.

\noindent\textbf{Text-to-triplane network} $T$ outputs a triplane representation from the text embedding $e$. $T$ is composed of two parts. The first part is a linear projection that maps the averaged text embedding from $\mathbb{R}^{C_e}$ to $\mathbb{R}^{3C_TH_TW_T}$, which is then reshaped to a triplane \cite{EG3D} representation $\mathbb{R}^{3 \times C_T \times H_T \times W_T}$. Note that $C_T$ is the number of triplane features and $H_T \times W_T$ denotes the height and width of each plane. The second part of $T$ is a text-conditioned, 3D-aware Triplane ConvNeXt network to enhance the triplane features.
We construct the network by stacking $N$ ConvNeXt blocks, where each block consists of four modules. 
The first is a multihead cross-attention module \cite{Transformer} between text embedding and triplane features. Each pixel of a plane is used as one query token, while each text token serves as the key and value. This cross-attention module is beneficial for higher generalizability and better quality especially when the dataset is large-scale (see \S\ref{sec:ablation}). 
The second module is a 3D-aware convolution borrowed from \cite{RODIN}. Due to the fact that each pixel $(i, j)$ in a plane can be associated with the whole column or row in the other two planes, this 3D-aware convolution is proposed to concatenate the features of $(i, j)$ with the averaged features of $(i, :)$ and $(:, j)$, and then perform a 2D convolution. 
The third is a depth-wise convolution with kernel size $7\times7$ to aggregate spatial information per channel, as inspired from ConvNeXt \cite{ConvNeXt}. 
The last is a feedforward network that is made up of two inverted linear layers to mix channel information. 
We perform convolutions and linear layers per plane, since we empirically find otherwise the information of non-correlated regions will be mixed that might slightly degrade the performance. 
The 3D-aware convolution and the feedforward network can be efficiently implemented by group convolutions using PyTorch. 
A residual connection is added to each module to avoid gradient vanishing. 
See \supp for the illustration of our text-to-triplane achitecture and the Triplane ConvNeXt block.

\noindent\textbf{Triplane-to-Mesh generator} $\nu$
generates a differential mesh from the triplane features. We use DMTet \cite{DMTet} as the mesh representation. DMTet represents a differential tetrahedral geometry as a signed distance field (SDF) defined on a deformable tetrahedral grid. The mesh vertices $V$ and their connections (mesh faces) are predefined on the initial grid. A deformation network is optimized to offset these predefined vertices for refined triangle meshes and finer details. The SDF values of the shifted points are learned by an SDF network to represent the final distance to the surface. The zero-level set of SDF values represents the surface of the triangle meshe. Moreover, a color network is leveraged to learn the color of each vertex. Overall, we employ three separate networks, \ie the SDF network, the deformation network, and the color network, to optimize DMTet. The input of each network is the concatenation of the triplane features and the positions with their sinusoidal encoding of the predefined mesh vertices.

\input{figures/pig}

\noindent\textbf{The inference stage of AToM}
 is a feedforward progress that gets a textured mesh directly from the text input, and is finished in less than one second without the need of optimization. During inference, once a text is given, AToM first gets the text embedding $e$ through the text encoder and next passes $e$ to the triplane generator to obtain the features. AToM then inputs the vertices of the DTMet grid to query the triplane features, encodes the vertices positions, and passes the concatenation of triplane features, positional encoding, and positions to the SDF, the deformation, and the color networks to output the textured mesh.

\subsection{Two-Stage Amortized Optimization}\label{sec:training}
Optimizing a text-to-mesh end-to-end is problematic due to the topology constraints \eg triangle connections, inherent in the differentiable mesh. Fantasia3D \cite{Fantasia3D} makes such a direct training of text-to-mesh possible, but requires sophisticated parameter tuning for each prompt. Refer to the rather poor geometry without per-prompt tuning in \supp. GET3D \cite{Get3D} also shows the possibility of training an unconditional mesh generator, but is limited to specific categories such as chair and car, and requires the ground-truth 3D data during training. 
We show that a trivial end-to-end training for AToM leads to divergent optimization (\S\ref{sec:ablation}). To address this unstable optimization, we propose a two-stage amortized optimization: a NeuS volumetric training as warmup followed by a differentiable mesh training.

\inlinesection{First stage: volumetric optimization}
We use volumetric rendering in the first stage to warmup the SDF network. In this stage, we use the NeuS \cite{NeuS} representation, and only optimize the triplane generator, the SDF network, and the color network. We render a low-resolution (\eg $64\times64$) image by accumulating points' colors along the ray. The obtained renderings are added noise and passed to the text-to-image diffusion prior to provide the guidance through SDS loss \cite{DreamFusion}. Mathematically, given a pixel in the rendered image, the ray emitted from this pixel is denoted as $\left\{\mathbf{p}_i=\mathbf{o}+t_i \mathbf{v} \mid i=\right.$ $\left.1, \ldots, n, t_i<t_{i+1}\right\}$, where $\mathbf{o}$ is the center of the camera, $\mathbf{v}$ is the unit direction vector of the ray, $n$ is the number of points per ray, $t$ denotes the distance from $\mathbf{o}$. NeuS volumetric rendering is given by:
\begin{equation}\label{eqn:neus}
\small
\begin{aligned}
    \hat{C} &= \sum_{i=1}^n T_i \alpha_i c_i, \quad T_i = \prod_{j=1}^{i-1}\left(1-\alpha_j\right) \\
    \alpha_i &= \max \left(\frac{\Phi_s\left(f\left(\mathbf{p}\left(t_i\right)\right)\right) - \Phi_s\left(f\left(\mathbf{p}\left(t_{i+1}\right)\right)\right)}{\Phi_s\left(f\left(\mathbf{p}\left(t_i\right)\right)\right)}, 0\right)
\end{aligned}
\end{equation}
where $\phi_s(x)=s e^{-s x} /\left(1+e^{-s x}\right)^2$ is the logistic density distribution. $s$, $f$, $c$ are the learnable parameter of NeuS, the SDF network, and the color for point $i$, respectively.

\inlinesection{Second stage: mesh optimization}
For the second stage, we use differentiable mesh representation. Fast and memory-efficient mesh rasterization is leveraged, allowing the system be trained with high-resolution renderings (\eg $512 \times 512$). The same SDS loss as the first stage is used as guidance. 
The deformation network initialized with zeros is included in the optimization, \ie the vertices of the initial mesh grid without offsets are used as query points. Since SDF and color networks are warmed up during the first stage, the main goal of the second stage is improve the quality of the geometry and texture through the high-resolution renders in mesh representation.


%% file: figures/pig.tex
\begin{figure*}[t]
\centering
\begin{subfigure}{0.49\textwidth}
  \centering
  \includegraphics[width=\textwidth]{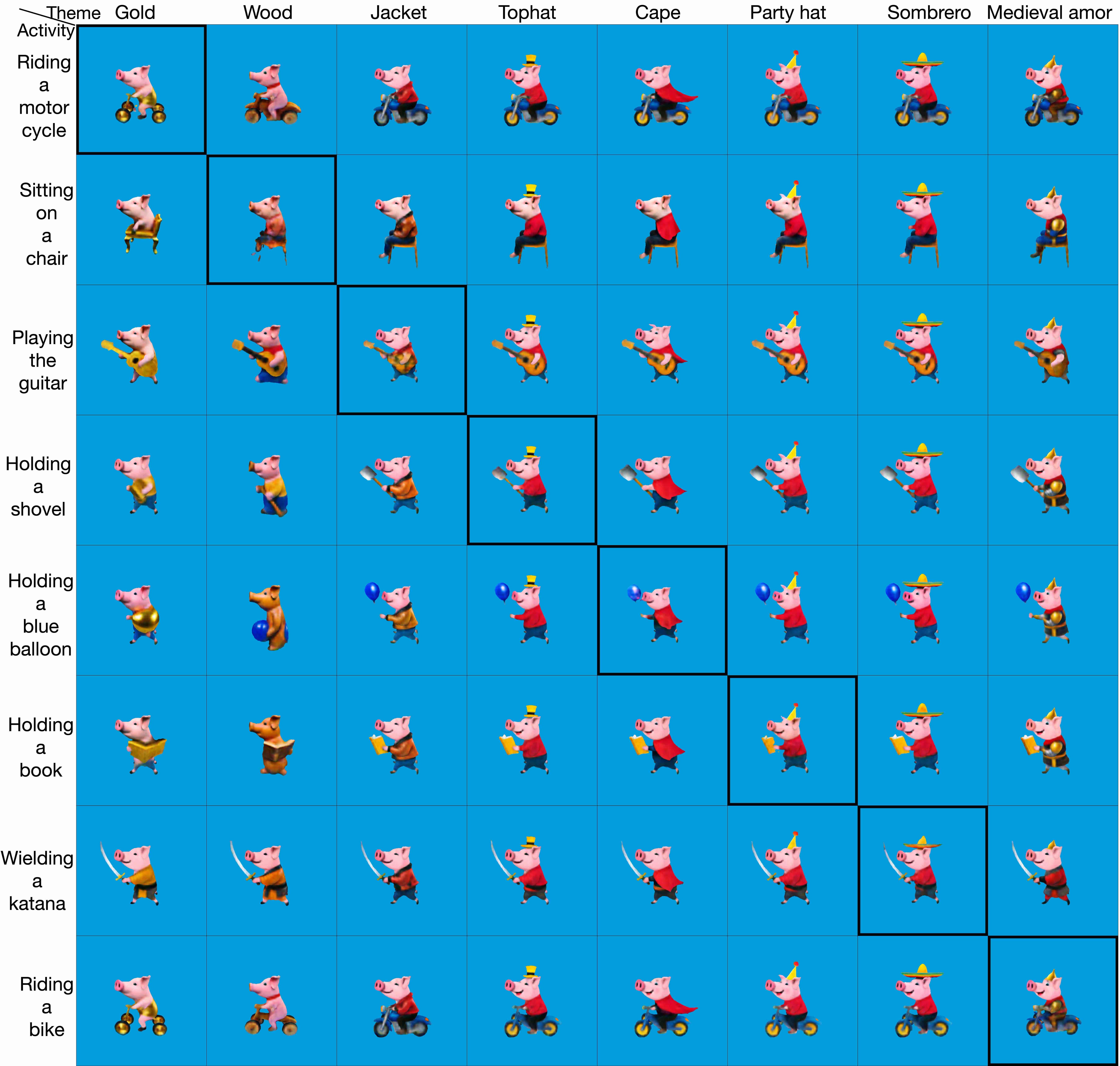}
  \caption{AToM}
\end{subfigure}%
\hfill
\begin{subfigure}{0.49\textwidth}
  \centering
  \includegraphics[width=\textwidth]{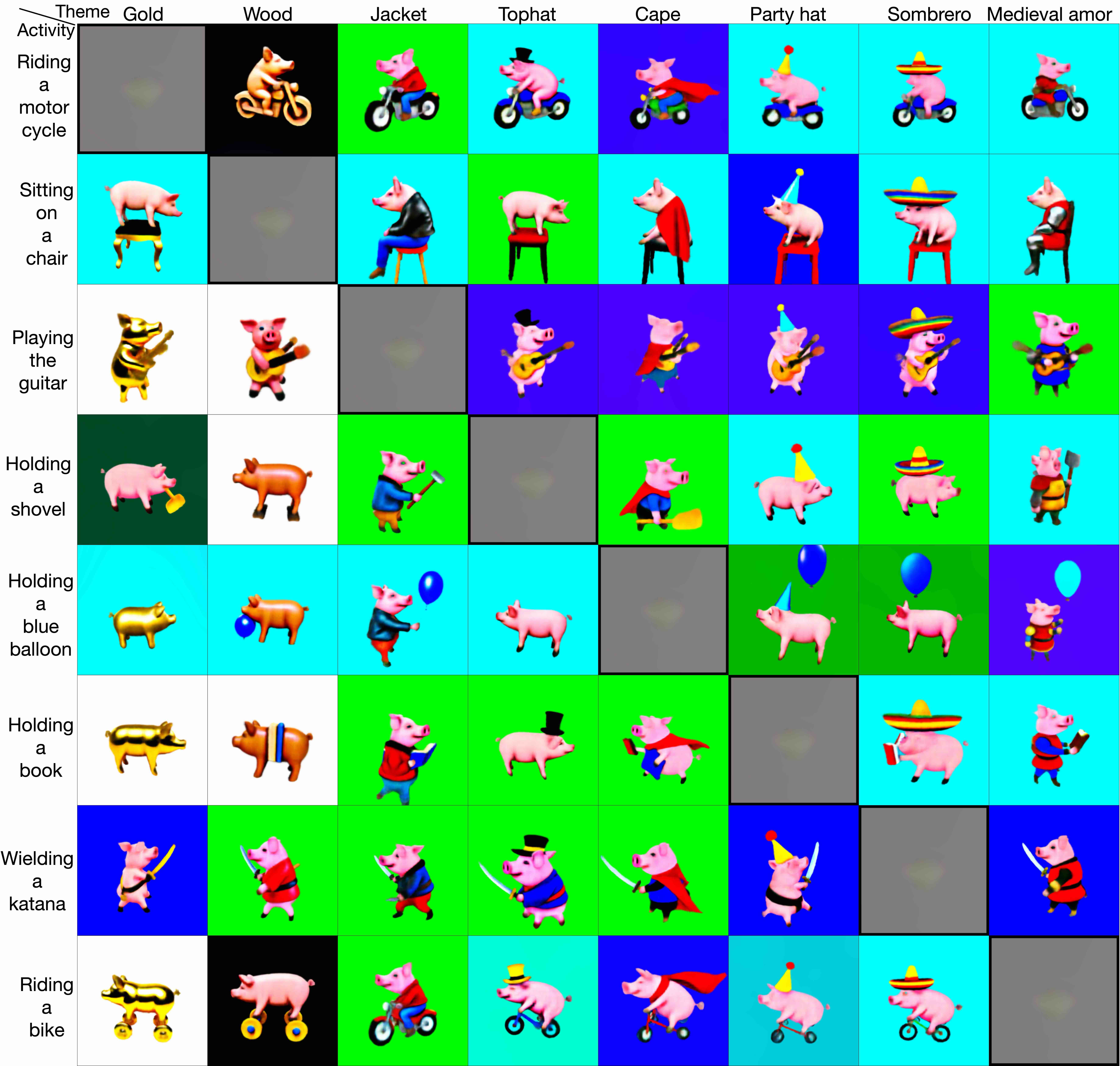}
  \caption{AToM Per-Prompt}
\end{subfigure}
\caption{\textbf{Comparing AToM to AToM Per-Prompt} on the Pig64 compositional prompt set (``a pig {activity} {theme}''), where each row and column represent a different activity and theme, respectively. The models are trained using 56 prompts and tested on all 64 prompts, while 8 unseen prompts are evaluated on the diagonal. As depicted in (a), AToM consistently generates pigs with a similar identity and a uniform orientation, indicating that AToM also promotes feature sharing across prompts, similar to ATT3D \cite{ATT3D}. Also, AToM generates 3D content with consistent quality, while per-prompt optimization cannot as shown in (b). Additionally, per-prompt optimization is more prone to overlooking certain details, such as the top hat in row 2 column 4 and the shovel in row 4 column 2 in (b), while AToM preserves them.
More importantly, AToM performs well on unseen prompts without further optimization, unlike the per-prompt solution.
}
\label{fig:pig}
\end{figure*}

%% file: sections/experiments.tex
\section{Experiment}\label{sec:exp}
We conduct comprehensive experiments on various benchmarks to show the effectiveness of AToM. We showcase our strong generalizability to unseen prompts, while the per-prompt solutions \cite{DreamFusion, TextMesh, Fantasia3D, Magic3D} cannot. We also demonstrate quantitatively and qualitatively that AToM outperforms ATT3D, the state-of-the-art amortized text-to-3D.  We show the capability of AToM in the large-scale benchmark while per-prompt solutions are prohibitive to train, and ATT3D produces indistinguishable results.

\input{tables/sota}

\inlinesection{Data}
We evaluate on two compositional datasets from ATT3D \cite{ATT3D}: Pig64 and Animal2400. Pig64 is structured according to the pattern ``a pig \{activity\} \{theme\}'' from 8 activities and $8$ themes. In total, there are $64$ prompts in Pig64, where $56$ are used as training and all prompts including $8$ unseen ones are used as testing. 
Animal2400 is constructed following the template ``\{animal\} \{activity\} \{theme\} \{hat\}''. There are 10 animals, 8 activities, 6 themes, and 5 hats. We experiment on the hardest spit ($12.5\%$ split), where only 300 prompts are used in training, and all 2400 prompts are tested including 2100 unseen ones. See \supp for dataset details.
We also evaluate on two datasets demonstrated in  DreamFusion~\cite{DreamFusion}: DF27, the $27$ prompts shown in their main paper and DF415, the $415$ prompts available on their project webpage.

\inlinesection{Evaluation Metrics}
 We employ the same evaluation metric, the \textit{CLIP R-probability}, as in ATT3D. The \textit{CLIP R-probability} gauges the average distance of the associated text with $4$ uniformly rendered views from the generated 3D model. This distance score indicates the level of confidence CLIP holds in regards to the relevance between the text prompt and the mutiple renderings from each 3D asset.

\inlinesection{Implementation Details}
We implement our AToM, reimplement ATT3D, and run per-prompt baselines \cite{DreamFusion,TextMesh,Magic3D,Fantasia3D} using the ThreeStudio \cite{threestudio} library. 
For all methods except Fantasia3D \cite{Fantasia3D} that requires using Latent Diffusion Model \cite{LDM}, we utlize Deep Floyd \cite{DeepFloyd} as the text-to-image prior, as it is found to offer higher quality across methods \cite{threestudio}. 
For text embedding, we use the same frozen T5 text encoder for prompt processing and text-to-triplane input. 
During the first stage, we render $64\times64$ resolution images with $64$ uniformly sampled points per ray. One can use $32$ points without significant difference. We optimize networks with learning rate 4e-4 and batch size 16 using 2 GPUs for 20K iterations on DF27, 4 GPUs for 10K iterations on Pig64, 8 GPUs for 100K iterations on DF415. 
For the second stage, we optimize with learning rate 2e-4 and batch size 16 using 4 GPUs for 10K iterations on DF27 and Pig64, and 8 GPUs for 50K iterations on DF415. See \supp for details.

\input{figures/gallery1}

\subsection{Unseen Interpolation Experiments}\label{sec:generalization}
As a significant benefit, \textit{AToM generalizes to interpolated prompts that are unseen during training}. This generalizability is not possessed by the per-prompt solutions. 
\figlabel\ref{fig:pig} showcases the differences of AToM compared to AToM per-prompt in the Pig64 compositional dataset. We highlight that  AToM per-prompt shares the same architecture but is trained in a per-prompt fashion. 
\textbf{We observe the following: }
\textbf{(1)} AToM can produce high-quality results of unseen prompts without further optimization, while per-promt optimziation cannot, as shown in the diagonal in \figlabel\ref{fig:pig}; 
\textbf{(2)} AToM generates pigs with a similar identity and a uniform orientation, which is not observed in per-prompt experiments, indicating that AToM promotes feature sharing across prompts;
\textbf{(3)} Per-prompt optimization is more prone to overlooking certain details, such as the top hat in row 2 column 4 and the shovel in row 4 column 2, due to the necessity for per-prompt parameter tuning, while AToM yilds a consistent quality across prompts. 
In \supp, we further illustrates the training dynamics of AToM compared to AToM per-prompt, AToM significantly outperforms its per-prompt version under the same training budgets. Trained both to convergence, we observe a reduction of training iterations by over 20 times of AToM \vs AToM per-prompt.
\supp also qualitatively compare AToM to ATT3D in Pig64 and the harder dataset Animal2400, where we again show the obvious improvements of AToM against ATT3D. Refer to \supp for details.

\input{figures/sota-df415}

\subsection{Compare to the State-of-the-Art}\label{sec:sota}
Tab. \ref{table:sota} presents our quantitative results in terms of CLIP R-probability on Pig64, DF27, and DF415 benchmarks, compared to the amortized text-to-NeRF method ATT3D, and per-prompt approaches \cite{DreamFusion, TextMesh, Fantasia3D, Magic3D}. In addition to reporting the official results, We also reproduce ATT3D using the same diffusion prior \cite{DeepFloyd} as AToM, denoted ATT3D-IF$^{\dagger}$ for a fair comparison. 
From the experiments, \textbf{one can observe the following:}
\textbf{(1)} AToM achieves a higher CLIP R-probability of $75.00\%$ than ATT3D ($64.29\%$) on Pig64's unseen prompts, indicating its stronger capability to generalize to unseen prompts. 
\textbf{(2)} Across the training (seen) prompts in Pig64 and DF27, AToM surpasses DreamFusion\cite{DreamFusion} and Fantasia3D \cite{Fantasia3D} on both datasets. In comparison to TextMesh \cite{TextMesh} and Magic3D \cite{Magic3D}, AToM slightly lags in CLIP R-probability in Pig64 seen prompts; however, visually, AToM exhibits more consistent results as shown in \figlabel\ref{fig:pig}. 
\textbf{(3)} Across all benchmarks shown in Tab. \ref{table:sota}, AToM showcases superior performance compared to ATT3D, highlighting AToM's effectiveness over ATT3D. Specifically, in DF415, AToM attains $81.93\%$ accuracy, much higher than ATT3D ($18.80\%$).

\figlabel\ref{fig:sota-df415}  show the qualitative comparison between AToM and ATT3D in the large-scale bechmark DF415. ATT3D mostly outputs indistinguishable 3D assets across various prompts. Conversely, AToM excels in managing large-scale benchmarks, handling complex prompts, and achieving consistently higher accuracy and higher quality than ATT3D. For qualitative comparisons against ATT3D and per-prompt solutions in Pig64 and DF27, see \supp. 
We observe AToM can achieve a comparable performance to the state-of-the-art with consistent quality across prompts unlike per-prompt solutions. 

%% file: tables/sota.tex
\begin{table}[t]
\centering
\caption{\textbf{Compare AToM to the state-of-the-art}. CLIP R-probability$\uparrow$ is reported. The per-prompt methods in seen prompts are \demph{deemphasized}. Per-prompt solutions have not been experimented in Animal2400 and DF415 due to their prohibitive computation. ATT3D's results are from the original paper \cite{ATT3D}. ATT3D-IF$^{\dagger}$ denotes our reproduced version using Deep Floyd \cite{DeepFloyd} as prior. 
}
\label{table:sota}
\resizebox{0.48\textwidth}{!}{%
\begin{tabular}{@{}c|ccc|c|cc}
Method/Dataset & Pig64 unseen & Pig64 seen & {Pig64 all} & Animal2400 &  {DF27} & {DF415} \\
\hline
{DreamFusion-IF} & 0 & \demph{0.7143} & 0.6250 & - & \demph{0.8889} & - \\
{TextMesh-IF} & 0 & \demph{0.8036} & 0.7031 & - & \demph{0.9259} & - \\
{Fantasia3D} & 0 & \demph{0.5357} & 0.4688 & - & \demph{0.7037} & - \\
{Magic3D-IF} & 0 & \demph{0.8036} & 0.7031 & - & \demph{0.8519} & - \\
\midrule
{ATT3D} & 0.3750 & 0.6071 & 0.5781 & 0.11 & 0.6296 & - \\
{ATT3D-IF$^{\dagger}$} & 0.6250 & 0.6429 & 0.6406 & 0.1671 & 0.8519 & 0.1880 \\
\textbf{AToM (\textbf{Ours})} & \textbf{0.7500} & 0.7500 & \textbf{0.7500} & \textbf{0.3442} & \textbf{0.9259} & \textbf{0.8193} \\
\end{tabular}
}
\end{table}


%% file: figures/gallery1.tex
\begin{figure*}[h]
\centering
\includegraphics[trim={0 0cm 0 0},clip,width=0.995\textwidth]{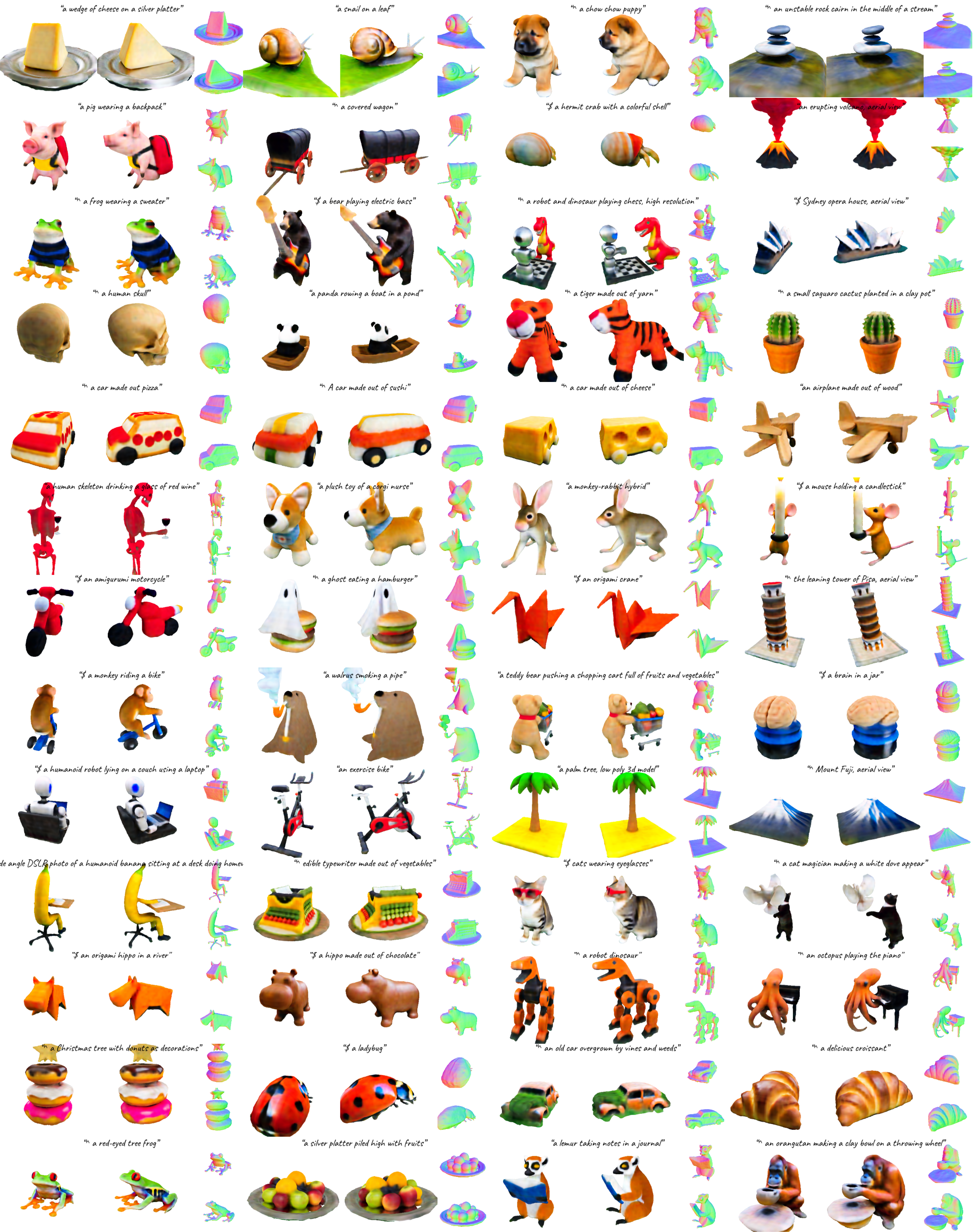}
\vspace{-1em}
\caption{\textbf{Gallery of AToM} evaluated in DF415. Here \^~ and \$ denote ``a zoomed out DSLR photo of'' and ``a DSLR photo of'', respectively.}
\label{fig:gallery}
\end{figure*}

%% file: figures/sota-df415.tex
\begin{figure*}[h]
\centering
\includegraphics[trim={0 0 0 0},clip,width=1.0\textwidth]{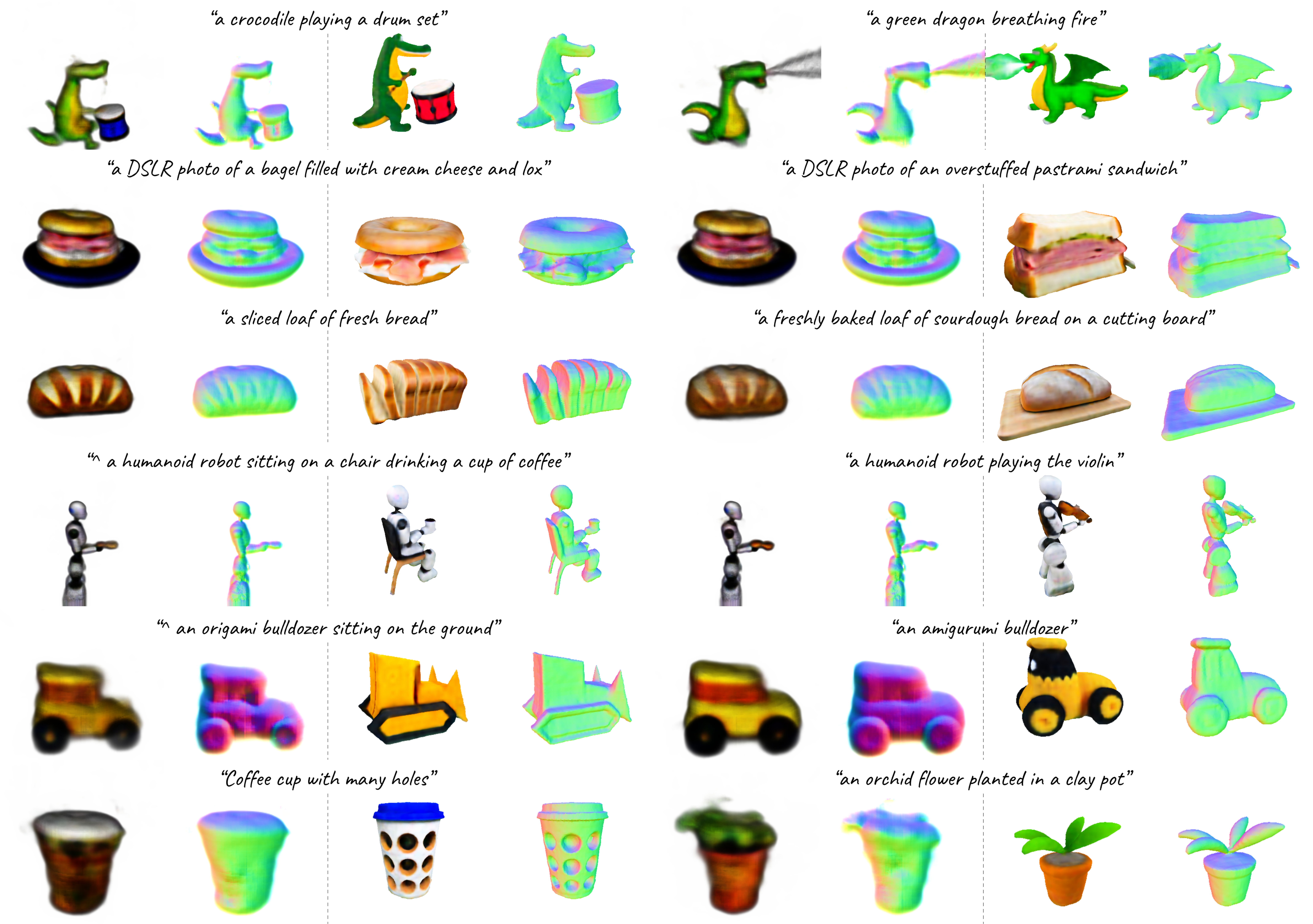}
ATT3D-IF$^{\dagger}$ \hspace{2cm} \textbf{AToM (Ours)} \hspace{3cm} ATT3D-IF$^{\dagger}$ \hspace{2cm} \textbf{AToM (Ours)} \\
\caption{\textbf{Compare AToM to ATT3D-IF$^{\dagger}$} evaluated in DF415. In each row, we mostly show results from two similar prompts. While ATT3D producing indistinguishable results for similar prompts, AToM handles the complexity of  prompts and achieves significantly higher quality than ATT3D. \^~ in the text denotes “a zoomed out DSLR photo of ”. One can also observe clear improvements of AToM over the original ATT3D by cross-referencing with their paper.}
\label{fig:sota-df415}
\end{figure*}

%% file: sections/ablation.tex
\section{Analysis and Ablation Study}\label{sec:ablation}

We perform ablation studies in DF415 in \figlabel\ref{fig:ablation}. 
We investigate: (1) the effect of two-stage training by comparing AToM with an end-to-end single-stage training version, (2) the effects of the second stage by comparing to the first-stage output, (3) the effect of triplane by comparing to AToM with HyperNet-InstantNGP (Hyper-INGP for short) used in ATT3D  as a replacement for positional encoding, (4) the effect of architecture designs including ConvNeXt, 3D-aware convolution, and cross attention.

\input{figures/ablation}

\inlinesection{Two-stage training significantly improves convergence}
Training AToM end-to-end using a single stage, \ie training solely by the second stage from scratch, leads to poor optimization. In DF415, we observe that the training diverges at a very early stage (in iteration 2000), generating black images without objects after that. We provides the results without two-stage training at iteration 2000 before the divergence in \figlabel\ref{fig:ablation} column 1. AToM single stage results in the lowest accuracy ($7.47\%$), significantly worse than AToM full ($81.93\%$).
\figlabel\ref{fig:ablation} demonstrates the coarse and noisy visual results due to its divergent training. We also ablate single-stage training in a smaller dataset DF27 provided in \supp, where the model trained through single stage can converge but still produces lower-quality results than the baseline for many prompts. Overall, these experiments clearly show the importance of two-stage training.

\inlinesection{Second-stage training yields meshes with improved visual quality}
\figlabel\ref{fig:ablation} column 2\&3 shows the usage of our second stage can slightly improves the accuracy. This is expected because the CLIP R-probability is used to estimate the correspondence between the generated 3D shape and the text prompt, for which, the first stage output is good enough. 
The higher-quality renderings from the second stage only have neglectable effect on the CLIP R-probability metric. 
Visually speaking, the second stage of AToM increases the rendering resolution, reduces noise, and enhances sharpness.

\inlinesection{Triplane \vs HyperNetworks}
We use text-to-triplane as a text-conditioned positional encoding instead of Hyper-INGP used in ATT3D. 
Hyper-INGP is a HyperNetworks-based \cite{HyperNetworks} positional encoding, which utilizes two linear layers to predict the weights for the Instant-NGP positional encoding network \cite{InstantNGP}. Hyper-INGP is not numerically stable and therefore requires special layers such as spectral normalization \cite{SpectralNorm} to stabilize optimization \cite{ATT3D}. 
Due to the difficulty of learning the weights of another neural network, Hyper-INGP shows low model capacity in the large-scale dataset, delivering a poor accuracy ($15.18\%$) in DF415 as indicated in \tablabel\ref{table:sota}. Its low capacity is also verified in our ablation study where we replace our text-to-triplane with Hyper-INGP in \figlabel\ref{fig:ablation} column 4: the accuracy of AToM first stage drops from $81.69\%$ to only $35.18\%$.  Visually speaking, less distinguishable results with Hyper-INGP are produced compared to our text-to-triplane network. To verify this performance drop is not due to the reduced complexity of the network, we also removed all ConvNeXt blocks in text-to-triplane and used two linear layers with spectral normalization to predict the features, which still significantly outperforms the Hyper-INGP counterpart, as indicated in  column 5 \vs column 4 ($77.11\%$ \vs $35.18\%$). We highlight that only difference between columns $4\&5$ is the positional encoding (Instant-NGP or triplane). These experiments clearly indicate the strength of our proposed text-to-triplane positional encoding.

\inlinesection{Triplane ConvNeXt designs}
In \figlabel\ref{fig:ablation} column 5 (w/o ConvNeXt), we experiment AToM without Triplane ConvNeXt. We observe an accuracy drop of $4.8$ points. 
The proposed Triplane ConvNeXt blocks are helpful in enhancing details, reducing noises, and more importantly, preserving components of complex prompts. 
We also tried to replace ConvNeXt blocks into Transformer blocks but found Transformer did not converge. We hypothesize that Transformer requires a significantly larger amount of data. 
We also perform ablation studies on the components of Triplane ConvNeXt blocks to investigate the effectiveness of 3D-aware convolution and cross attention, and reach lower CLIP R-Prob $79.76\%$ and $79.28\%$, respectively. These indicate that both 3D-aware convolution and cross attention improve the performance.

%% file: figures/ablation.tex
\begin{figure}[t]
\centering
\includegraphics[page=2,width=0.5\textwidth, trim= 0 4.31cm 9cm 0, clip]{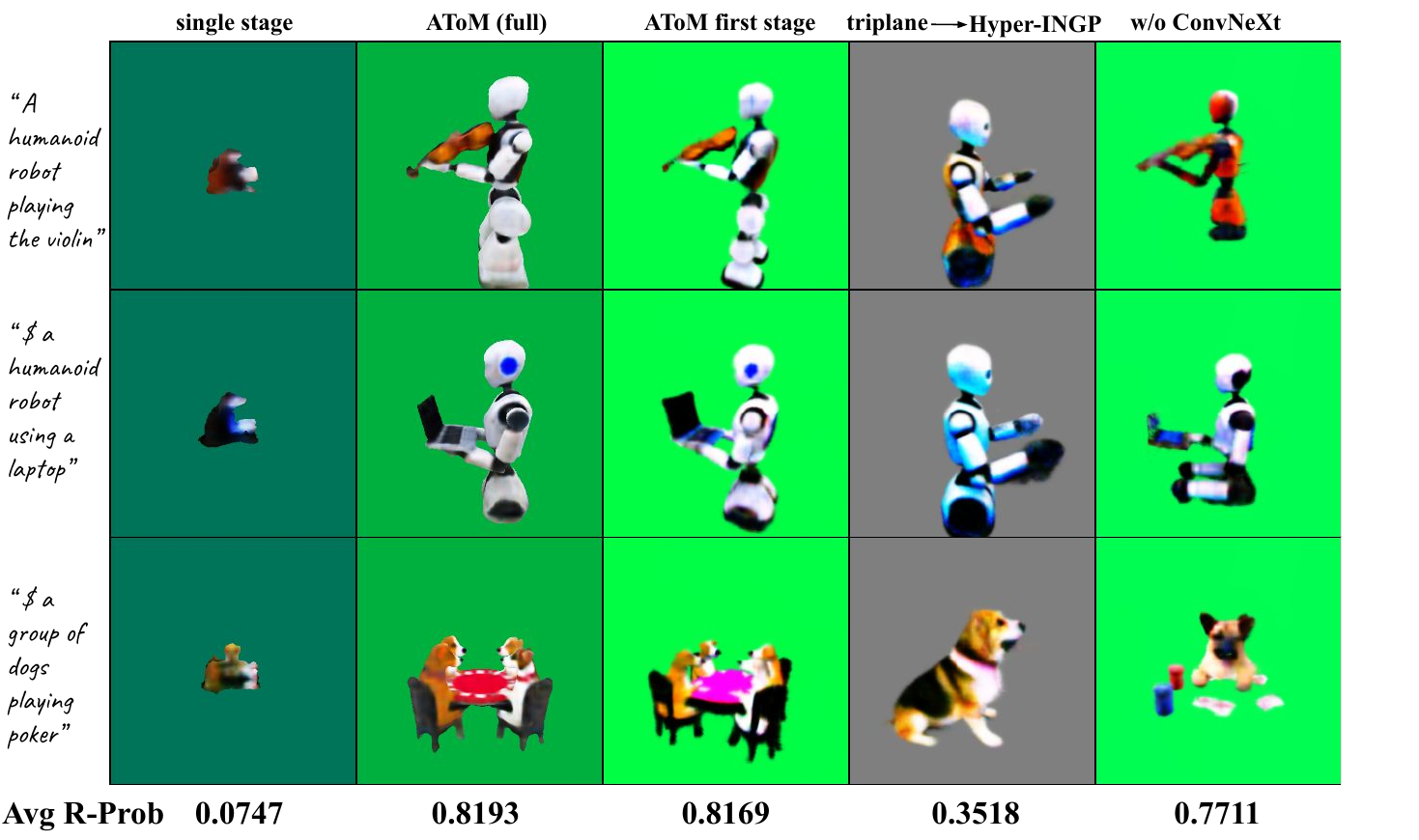}
\caption{\textbf{Ablation study.} We compare AToM full pipeline in column 2 against the end-to-end approach without two-stage training in column 1, the first-stage output without second-stage refinement in column 3, AToM first stage without triplane but employing Hyper-INGP used in ATT3D \cite{ATT3D} in column 4, AToM first stage without ConvNeXt blocks but using two linear layers with spectral normalization for text-to-triplane  in column 5. Quantitative results in average R-Probability evaluated in the entire DF415 dataset are provided at the bottom. \$~in the text denotes ``a DSLR photo of ''.
}
\label{fig:ablation}
\end{figure}

%% file: sections/conclusion.tex
\section{Conclusion}
This work proposes AToM, the first amortized text-to-mesh framework.
AToM introduces a 3D-aware text-to-triplane network, which leads to superior quality compared to the HyperNetworks counterpart used in ATT3D. AToM also adopts a two-stage amortized optimization to stabilize the text-to-mesh generation. AToM significantly reduces training time compared to per-prompt solutions due to geometry sharing of amortized optimization. 
More importantly, AToM demonstrates strong generalizability, producing high-quality 3D content for unseen prompts without further optimization. 
Compared to ATT3D, AToM achieves an accuracy more than $4$ times higher in DF415. Qualitatively, AToM outperforms ATT3D by providing distinguishable 3D assets with better geometry and texture. 
We believe AToM, along with the code, the pretrained models, and the generated 3D assets that will be made publicly available, will contribute to push the boundaries of text-to-mesh generation.



%% file: sections/supplement.tex
\renewcommand{\thesection}{\Alph{section}}
\renewcommand{\thetable}{\Roman{table}}
\renewcommand{\thefigure}{\Roman{figure}}
\setcounter{section}{0}
\setcounter{table}{0}
\setcounter{figure}{0}

\twocolumn[{%
\centering
  \includegraphics[height=1em]{figures/src/logo.png}
{\large\bf AToM: Amortized Text-to-Mesh using 2D Diffusion
}\\
\vspace{1em}
{\large\bf--- Appendix ---}
\vspace{2em}
}]



\section{Implementation Details}\label{sec:implementation}
\subsection{AToM implementation details}

AToM uses a similar camera and rendering setup to TextMesh \cite{TextMesh} in the first stage and similar to Magic3D \cite{Magic3D} in the second stage. We bound the objects in $2$ meters and set the camere $3$ meters away from the object center. We employ a field of view ranging from $40$ to $70$ in the first stage and from $30$ to $40$ in the second stage. Soft rendering with $50\%/50\%$ probabilities for textureless/diffuse shading is used to avoid learning flat geometry. 
We implement SDF, deformation, and color networks using three separate three-layer MLPs with hidden channels $64$. We empirically find that these separate networks slightly improve the generation quality than the single model used in ATT3D. 
Original texts without direction are used as input for the text-to-mesh network, while directional prompts with ``, front view'', ``, side view'', ``, back view'', ``, overhead view'' are used as text condition in the diffusion model. 
We utilize Deep Floyd \cite{DeepFloyd} as the text-to-image prior with guidance scale 20. A random noise from (0.2, 0.98) and (0.02, 0.5)

\subsection{ATT3D reimplementationdetails}
In the main paper, we report the quantitative results of the original ATT3D and our re-implemented version ATT3D-IF$^{\dagger}$.

\noindent\textbf{Original ATT3D} is not yet released. We retrieve the quantitative results from Figure 6 in their original paper. We compare with the original ATT3D in Table 1 in our manuscript.

\noindent\textbf{ATT3D-IF$^{\dagger}$} denotes our reimplementation using the exact same architecture, training parameters, and camera setups as mentioned in the original paper, except for the unavaiable ones (where we use the same as AToM). The only difference of ATT3D-IF$^{\dagger}$ from the original ATT3D is the diffusion prior: while the original ATT3D used their internal version, we adopt the publicly available IF model from Deep Floyd \cite{DeepFloyd}. We cannot achieve exact the same performance as ATT3D mostly due to the different diffusion prior.

\input{figures/triplane}


\input{figures/pig_two_stage}

\input{figures/pig_training}

\section{Method Details}

\noindent\textbf{Method Comparison.}
AToM is trained on many prompts simultaneously through SDS loss without any 3D data and generates textured meshes in less than 1 second during inference. This differs AToM from previous 3D reconstruction models such as GET3D \cite{Get3D}, which requires 3D Ground Truth and is thus limited by the availability of 3D data and the low diversity of 3D shapes.
Compared to Magic3D \cite{Magic3D}, the well-known per-prompt text-to-mesh work, we are similar in two-stage training, but differ from each other. Magic3D uses the results of the first stage to initialize the SDF parameters of the second stage through an optimization progress \cite{Magic3D}, leading to inaccurate initialization and cannot be trained amortized. 
Conversely, the AToM network remains unchanged across both stages. The first stage training in AToM serves as a warmup phase for the second stage. This approach uses the same SDF and color networks in both stages and eliminates the need for optimization of the SDF parameters, unlike Magic3D.
Last but not least, AToM differs from ATT3D \cite{ATT3D} in two important aspects:
(1) AToM is the first to enable amortized training for text-to-mesh generation, while ATT3D only supports text-to-NeRF; 
(2) AToM uses triplane to condition the generation of 3D content, which is more robust to training parameters and is more stable in optimization compared to the HyperNet-based solutions in ATT3D.

\noindent\textbf{Triplane ConvNeXt.}
We provides a pseudo code for the proposed Triplane ConvNeXt in Algorithm \ref{alg:triplane}. We illustrate Triplane ConvNeXt in \figlabel\ref{fig:triplane}. 

\begin{algorithm}[!tb]
\small
\caption{\small Code for Triplane ConvNeXt (PyTorch \cite{Paszke2019Pytorch} like)}
\begin{PythonA}[frame=none]
import torch.nn.functional as F
def forward(text_emb):
    # map averaged text_emb to triplane
    avg_emb = text_emb.mean(dim=1,keepdims=False)
    x = self.linear(avg_emb)
    # reshape to triplane 
    x = x.reshape(-1, self.c, self.h, self.w)
    # Triplane ConvNeXt blocks
    for i in range(self.num_blocks):
        inp = x
        # cross attention
        x = x + self.crossatt(x,text_emb)
        # 3D aware convolution
        x = x + F.relu(self.aware3dconv(x))
        # FeedForard network
        x = x + self.ffn(x)
        # residual connection
        x = inp + x
\end{PythonA}
\label{alg:triplane}
\end{algorithm}

\section{Additional Results}\label{sec:appendix_qualitative}
\subsection{Pig64}

\noindent\textbf{Dataset details.} 
Pig64 is structured according to the pattern ``a pig \{activity\} \{theme\}'' from 8 activities and $8$ themes. 

activities = [``riding a motorcycle'', ``sitting on a chair'', ``playing the guitar'', ``holding a shovel'', ``holding a blue balloon'', ``holding a book'', ``wielding a katana'', ``riding a bike''].

themes = [``made out of gold'', ``carved out of wood'', ``wearing a leather jacket'', ``wearing a tophat'', ``wearing a cape'',  ``wearing a party hat'', ``wearing a sombrero'', ``wearing medieval armor'']

\noindent\textbf{Two stages of AToM on Pig64.} \figlabel\ref{fig:pig_two_stage} shows the comparisons of AToM first-stage outputs and AToM second-stage outputs. The mesh refinement stage (second stage) turns the NeuS representation to a high-resolution mesh representation and sharply increases visual quality. 

\noindent\textbf{Training dynamics.}
\figlabel\ref{fig:training} shows the training dynamics of AToM compared to AToM Per-prompt (per-prompt optimized versions of AToM network). 
Amortized training significantly reduces training cost per prompt. While per-prompt optimization typically requires 2000 - 8000 training batches, amortized optimization with AToM reaches the same-level accuracy with only 142 training batches per-prompt. In other words, AToM reduces the training time in this compositional dataset by more than $10\times$.

\input{figures/pig_att3d}

\noindent\textbf{ATT3D reimplementation.}
We compare our reimplemented ATT3D-IF$^{\dagger}$ to the original ATT3D in \figlabel\ref{fig:pig_att3d}.

\subsection{Animal2400}

\noindent\textbf{Dataset details.} 
We also include comparisons of AToM to ATT3D on Animal2400 $12.5\%$ split, where only 300 prompts are used in training and all 2400 prompts are used in testing. 
Animal2400 is constructed following the template ``\{animal\} \{activity\} \{theme\} \{hat\}''. There are 10 animals, 8 activities, 6 themes, and 5 hats.

animals = [``a squirrel'', ``a raccoon'', ``a pig'', ``a monkey'', ``a robot'', ``a lion'', ``a rabbit'', ``a tiger'', ``an orangutan'', ``a bear'']

activities = [``riding a motorcycle'', ``sitting on a chair'', ``playing the guitar'', ``holding a shovel'', ``holding a blue balloon'', ``holding a book'', ``wielding a katana'', ``riding a bike'']

themes = [``wearing a leather jacket'', ``wearing a sweater'', ``wearing a cape'', ``wearing medieval armor'', ``wearing a backpack'', ``wearing a suit'']

hats = [``wearing a party hat'', ``wearing a sombrero'', ``wearing a helmet'', ``wearing a tophat'', ``wearing a baseball cap'']

\noindent\textbf{Results.}
AToM significantly outperforms ATT3D-IF$^{\dagger}$ as shown in \figlabel\ref{fig:animal2400}. 
Quantitatively, AToM achieves 0.3422 CLIP R-Probability, higher than the original ATT3D (0.11) and ATT3D-IF$^{\dagger}$ (0.1671). 
AToM trained in this $12.5\%$ split seems even outperforms the original ATT3D trained in $50\%$ split by cross referencing Fig. 8 in ATT3D \cite{ATT3D}.

\input{figures/animal2400}

\input{figures/singlestage}

\input{figures/sota}
\subsection{DF27}
We compare AToM with per-prompt solutions and ATT3D-IF$^{\dagger}$ in \figlabel\ref{fig:sota}. Note that we do not expect the performance of amortized training to be better than per-prompt training for the seen prompts.

\section{Additional Ablation Study}\label{sec:appendix_ablation}
\noindent\textbf{Single-stage training in DF27}. In manuscript, we show that single-stage training in DF400 leads to divergent optimzation. Here, we further demonstrate that single-stage training can converge in smaller dataset (DF27), but still suffers from poor geometry. See examples in \figlabel\ref{fig:single_stge_df27}.

\section{Limitation and Future Work}
First, the quality of AToM is bounded by the diffusion prior employed. Throughout the work, we utilized IF-stage1 \cite{DeepFloyd} as the diffusion prior, which limits the high-frequency details due to its low-resolution input. The use of a higher-resolution diffusion like Stable Diffusion \cite{LDM} (SD) and IF-stage2 might improve the quality of AToM further. We made initial attempts to use SD, SD's variants including VSD \cite{ProlificDreamer} and MVDream \cite{MVDream}, and IF-stage2 but resulted in worse performance. We hypothesize the lower quality stems from the difficulties of their optimization in the amortized settings.
We believe how to correctly utilize stronger diffusion for amortized text-to-mesh is promising. 
Second, AToM used DMTet with SDF as the mesh representation, which is not capable of modeling surfaces of nonzero genus. More advanced mesh representation can be adopted to address this limitation, which is orthogonal to our study. 
Third, Janus problem also exists in some results of AToM, despite the fact that AToM alleviate it a lot mostly by geometry sharing in amortized optimization. We also tried progressive views, debiasing scores \cite{hong2023debiasing}, Perp-Neg \cite{Perp-Neg}, but empirically found that they did not  work trivially in the amortized setting. 
We leave these limitations and their potential solutions as future work.

%% file: figures/triplane.tex
\begin{figure}[tb]
\centering
\includegraphics[page=2,width=1.0\columnwidth, trim= 0 14.7cm 21cm 0, clip]{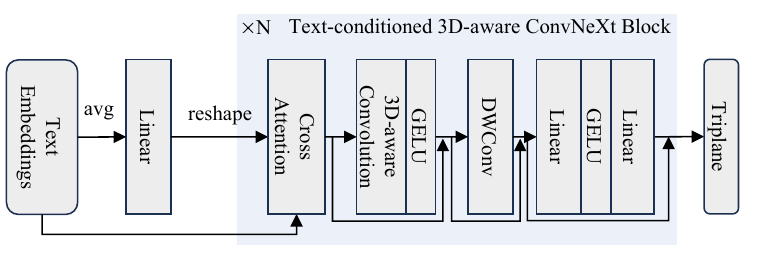}
\caption{\textbf{Text-to-Triplane architecture.} Triplane is generated from the averaged text embedding followed by a linear projection and then refined by multiple text-conditioned 3D-aware ConvNeXt blocks.}
\label{fig:triplane}
\end{figure}

%% file: figures/pig_two_stage.tex
\begin{figure*}[t]
\centering
\begin{subfigure}{0.49\textwidth}
  \centering
  \includegraphics[width=\textwidth]{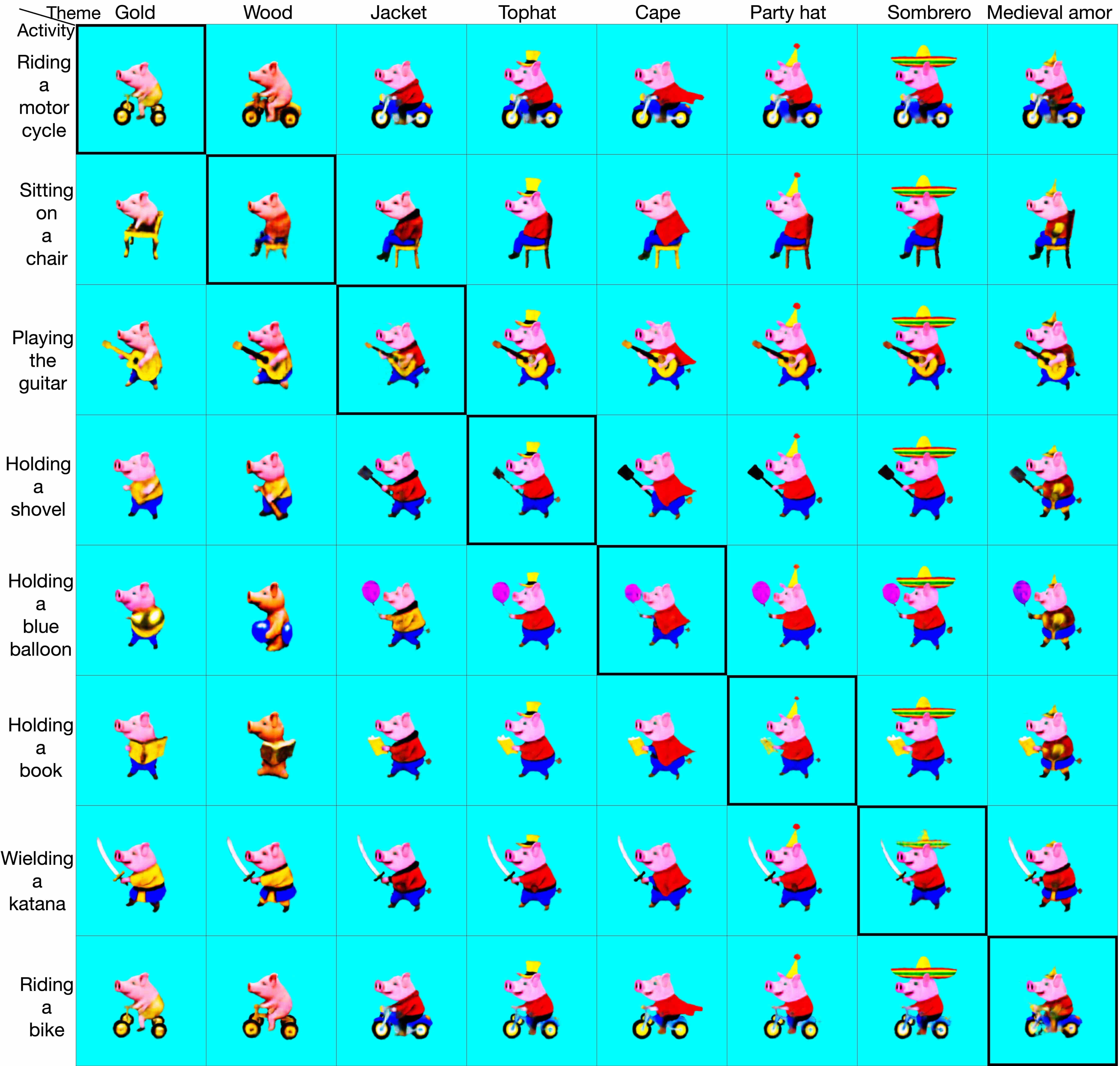}
  \caption{AToM Stage 1}
\end{subfigure}%
\hfill
\begin{subfigure}{0.49\textwidth}
  \centering
  \includegraphics[width=\textwidth]{figures/src/pig64/atom-stage2-pig.jpg}
  \caption{AToM Stage 2}
\end{subfigure}
\caption{\textbf{Results of AToM first stage (left) and second stage (right)} on the Pig64 compositional prompt set. The mesh refinement stage (second stage) turns the NeuS representation to a high-resolution mesh representation and sharply increases visual quality. }
\label{fig:pig_two_stage}
\end{figure*}

%% file: figures/pig_training.tex
\begin{figure}[t]
\centering
\includegraphics[width=\columnwidth]{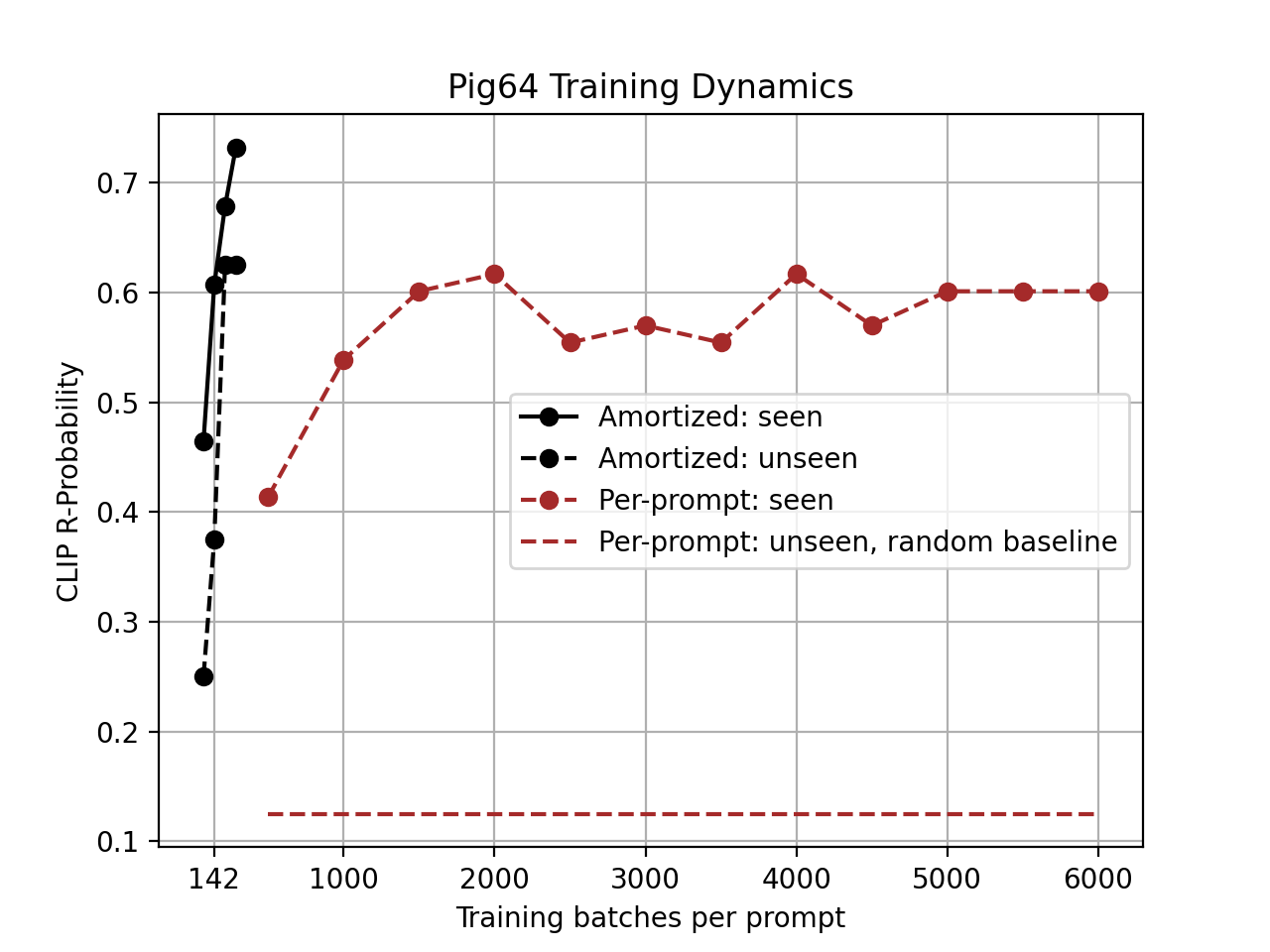}
\caption{\textbf{Training dynamics comparisons between AToM and AToM Per-prompt.}
Amortized training significantly reduces training cost per-prompt.
}
\label{fig:training}
\end{figure}

%% file: figures/pig_att3d.tex
\begin{figure*}[t]
\centering
\begin{subfigure}{0.49\textwidth}
  \centering
  \includegraphics[width=\textwidth]{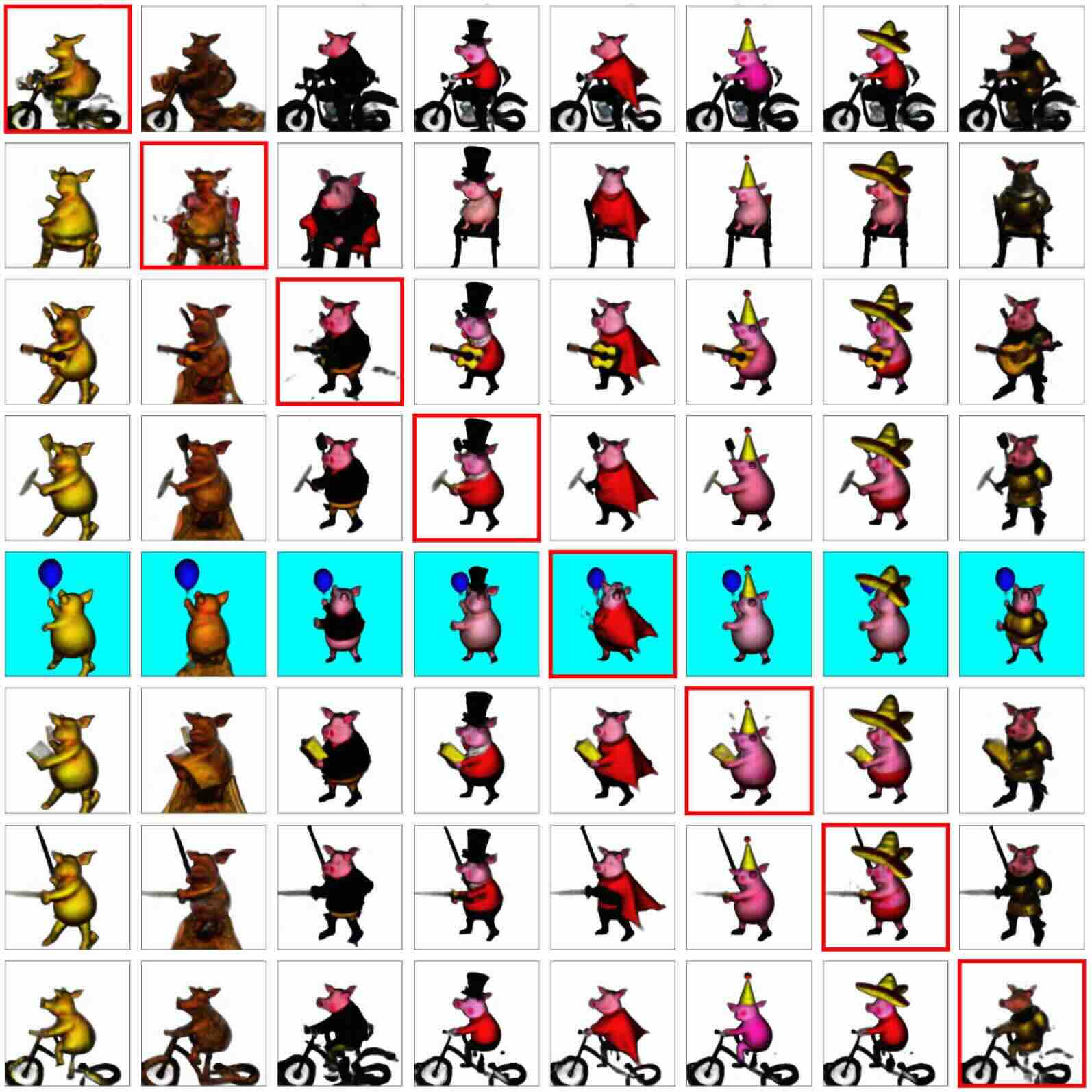}
  \caption{Original ATT3D}
\end{subfigure}%
\hfill
\begin{subfigure}{0.49\textwidth}
  \centering
  \includegraphics[width=\textwidth]{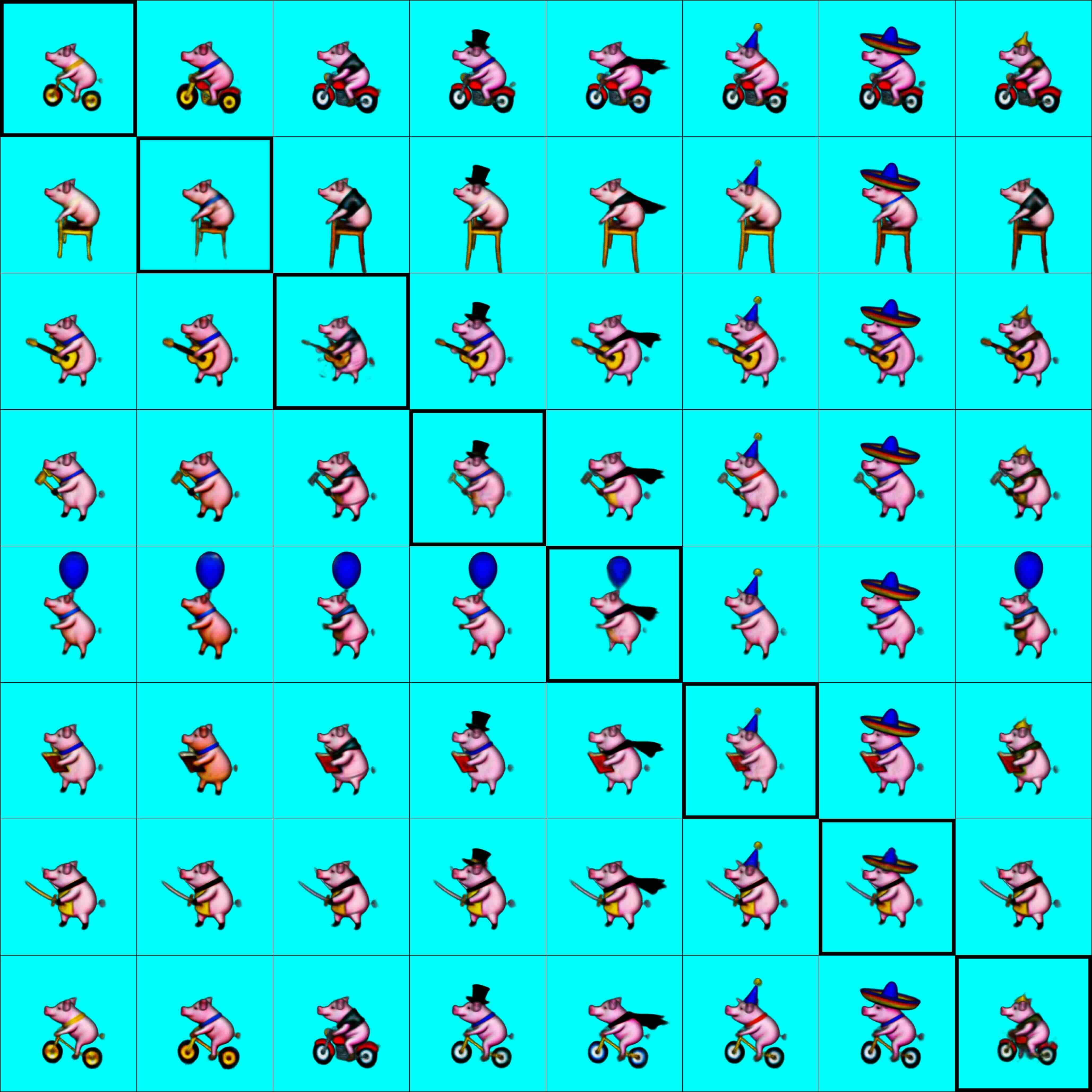}
  \caption{ATT3D-IF$^{\dagger}$}
\end{subfigure}
\caption{\textbf{Compare our reproduced ATT3D-IF$^{\dagger}$ at right to original ATT3D \cite{ATT3D} at left.}
Due to the distinct diffusion employed in ATT3D-IF$^{\dagger}$, the disparate outcomes from original ATT3D are expected. As strength, our reimplementation using Deep Floyd guidance facilitates more geometry sharing, yields results with less variances, and reduces noises. Especially, we highlight the unseen examples in the diagonal, where ATT3D-IF$^{\dagger}$ shows better generalizability then origianl ATT3D. As drawbacks, our reimplementation handles prompts sometimes worse than the original version, \eg not all pigs are made out of wood in the second column. Desipte the different implementation, AToM outperforms both versions, see \figlabel\ref{fig:pig_two_stage} for qualitative improvement and \tablabel\ref{table:sota} in main paper for quantitative improvements. 
}
\label{fig:pig_att3d}
\end{figure*}

%% file: figures/animal2400.tex
\begin{figure*}[t]
\centering
\begin{subfigure}{0.85\textwidth}
  \centering
  \includegraphics[width=\textwidth]{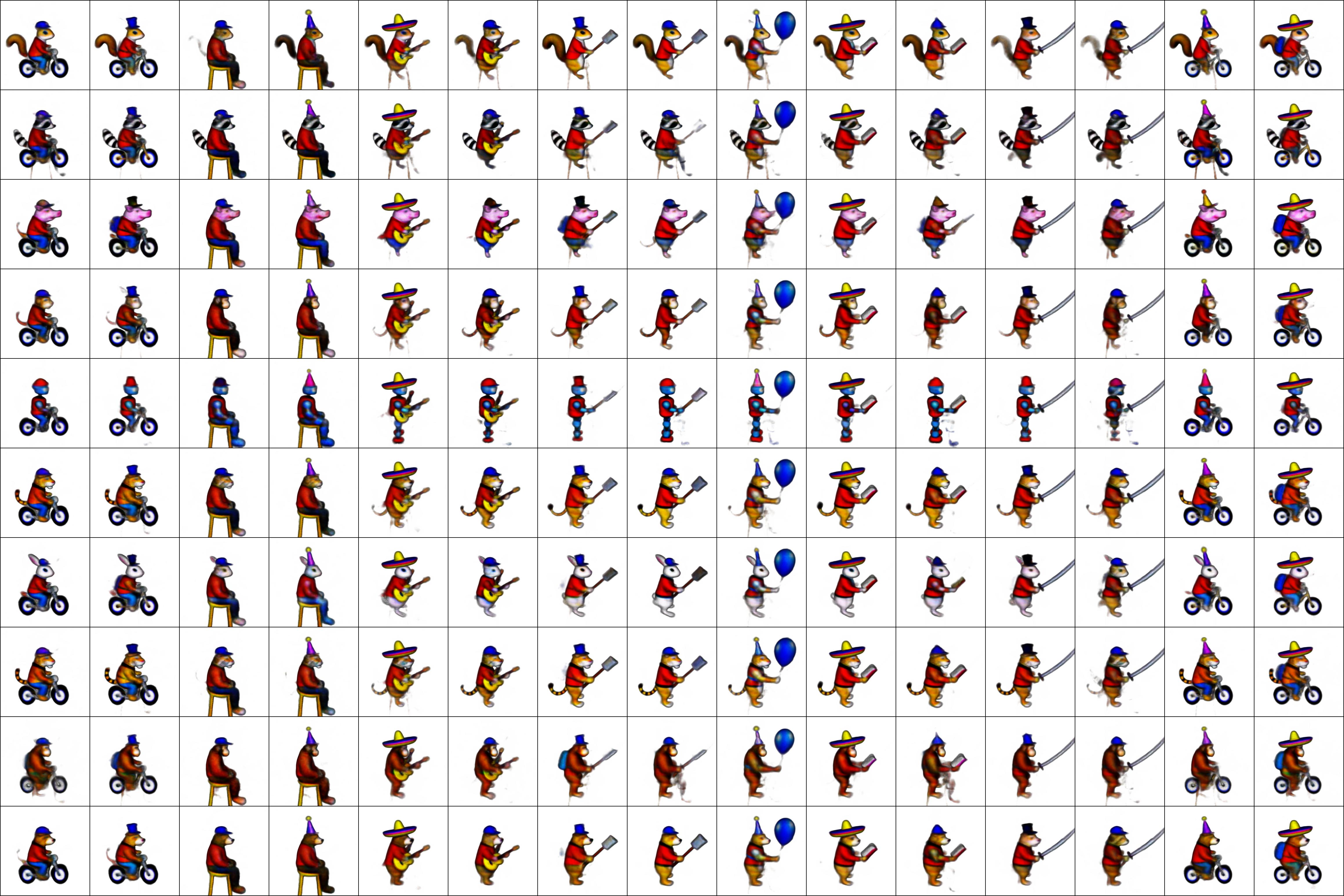}
  \caption{ATT3D-IF$^{\dagger}$. CLIP R-Prob in 2400 prompts: 0.1671.}
\end{subfigure}%
\hfill
\begin{subfigure}{0.85\textwidth}
  \centering
  \includegraphics[width=\textwidth]{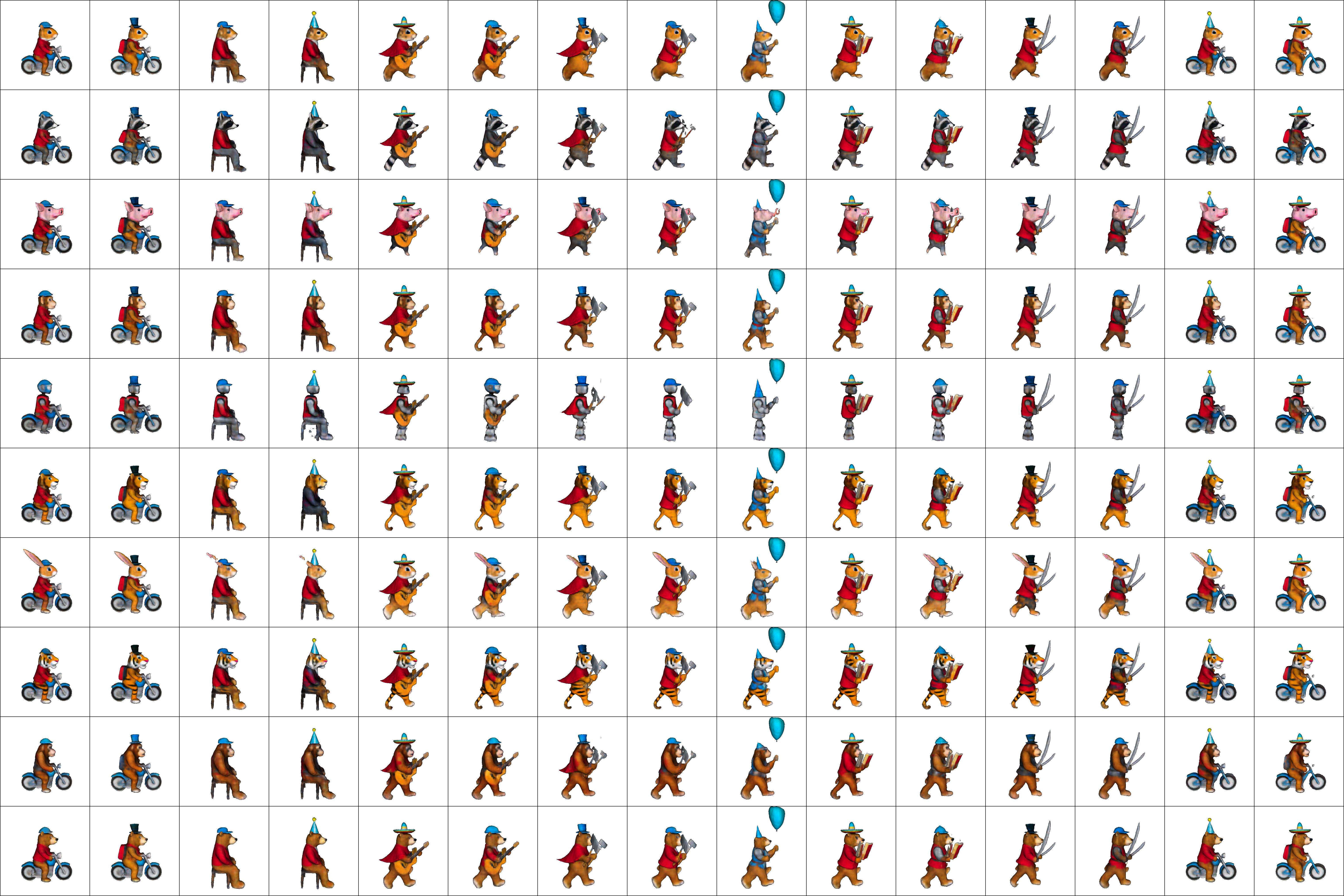}
  \caption{AToM. CLIP R-Prob in 2400 prompts: 0.3422.}
\end{subfigure}
\caption{\textbf{Compare AToM to ATT3D-IF$^{\dagger}$} on Animal2400 $12.5\%$ split. Trained only in 300 prompts, AToM also generalizes to all 2400 prompts, and significantly outperforms ATT3D and ATT3D-IF$^{\dagger}$.
See the overall improved quality and how AToM perserves the prompts when ATT3D-IF$^{\dagger}$ overlooks (\eg, the backpacks in the second column). }
\label{fig:animal2400}
\end{figure*}

%% file: figures/singlestage.tex
\begin{figure}[t]
\centering
\includegraphics[page=3,width=0.5\textwidth, trim= 0 7.2cm 12cm 0, clip]{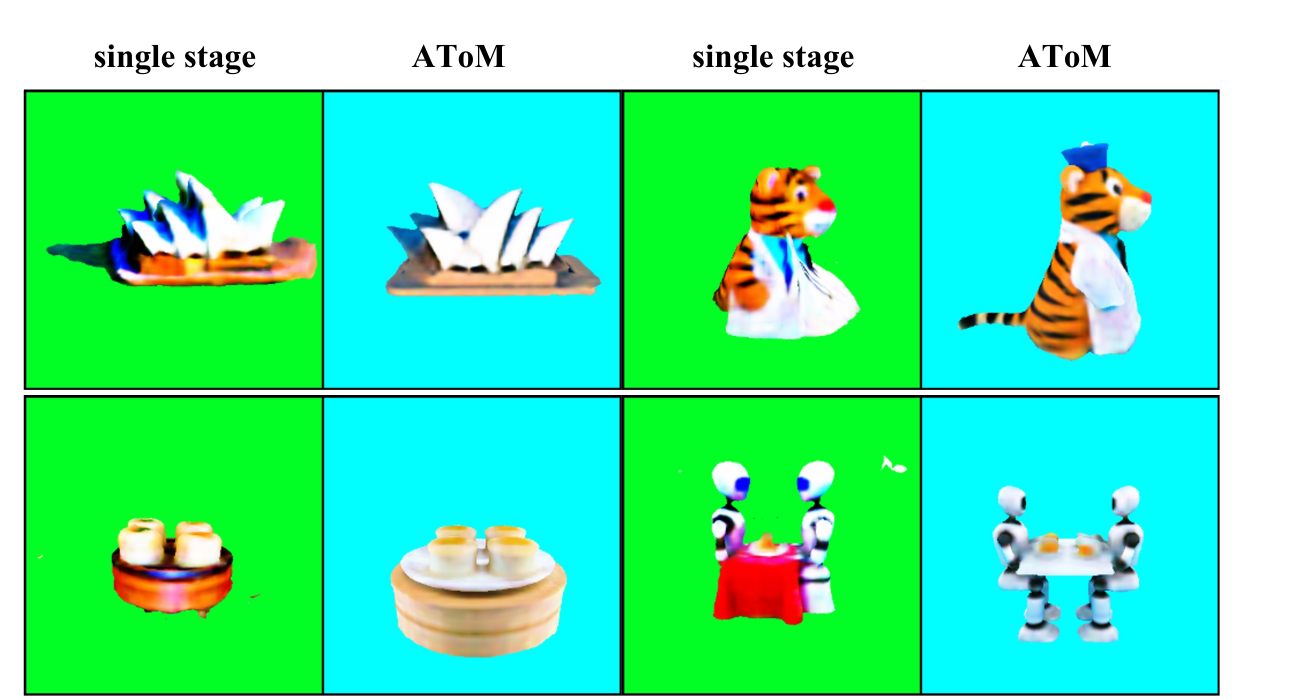}
\caption{\textbf{Compare Single-Stage AToM with AToM Full.} Single-stage training can converge in smaller dataset (DF27), but still suffers from poor geometry, compared to the two-stage training of AToM.
}
\label{fig:single_stge_df27}
\end{figure}

%% file: figures/sota.tex
\begin{figure*}[h]
\includegraphics[width=1.0\textwidth]{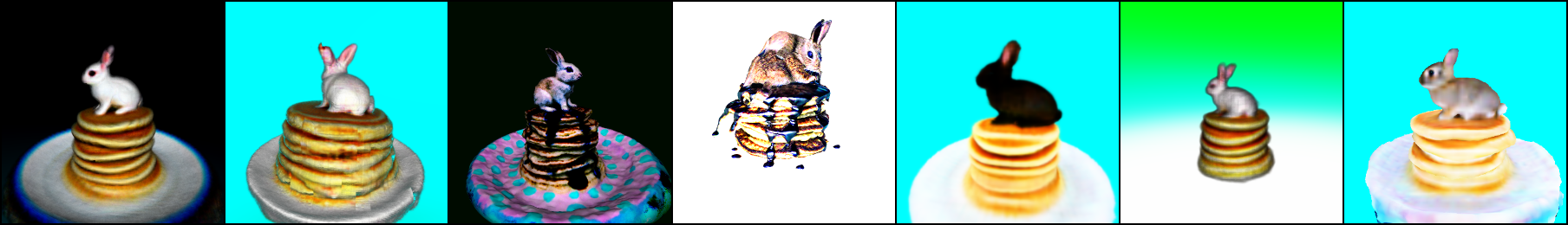}
\includegraphics[width=1.0\textwidth]{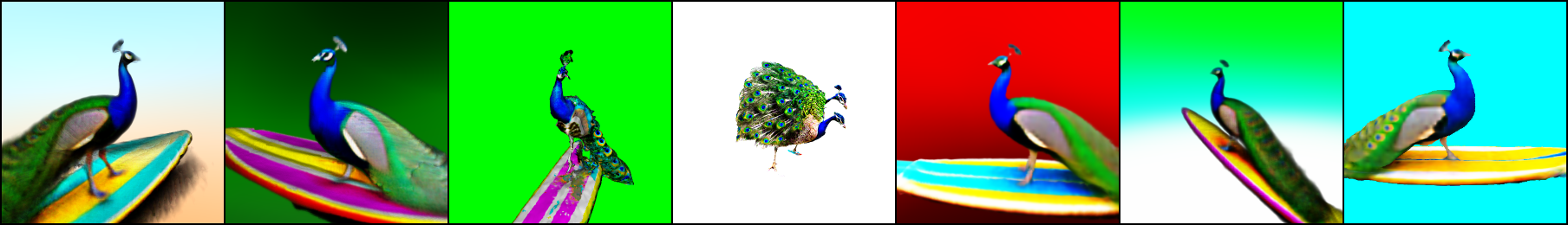}
\includegraphics[width=1.0\textwidth]{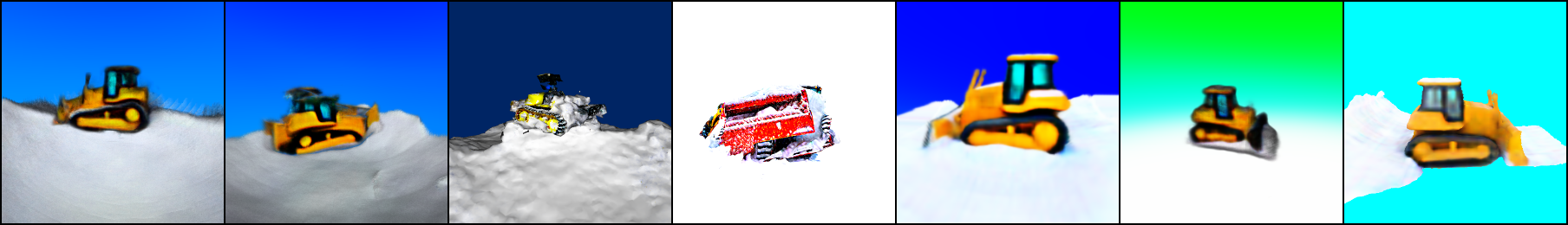}
\includegraphics[width=1.0\textwidth]{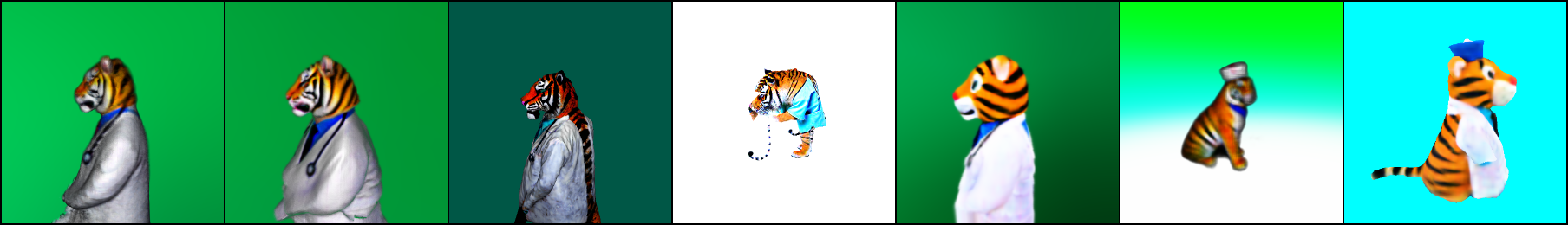}
\includegraphics[width=1.0\textwidth]{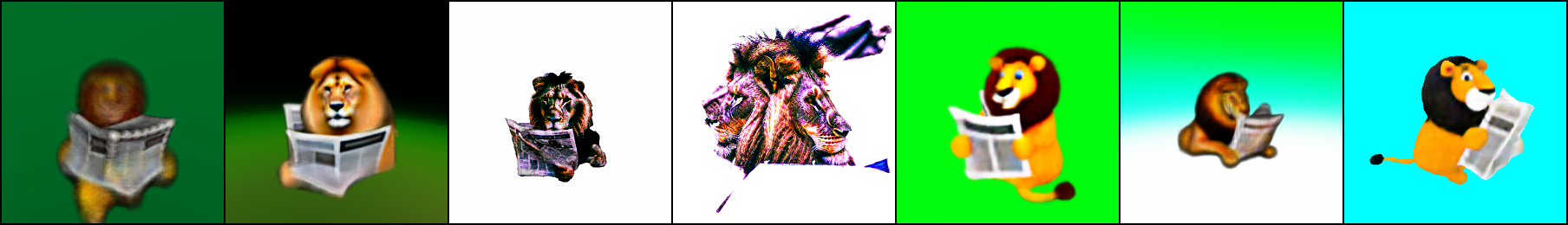}
\includegraphics[width=1.0\textwidth]{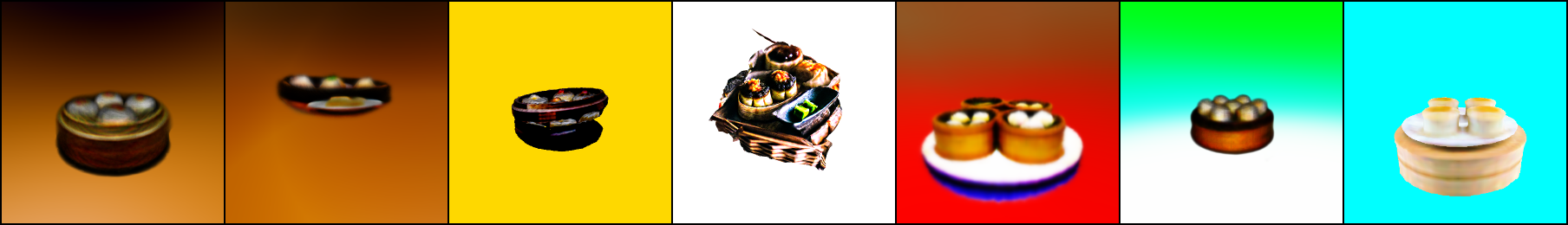}
\includegraphics[width=1.0\textwidth]{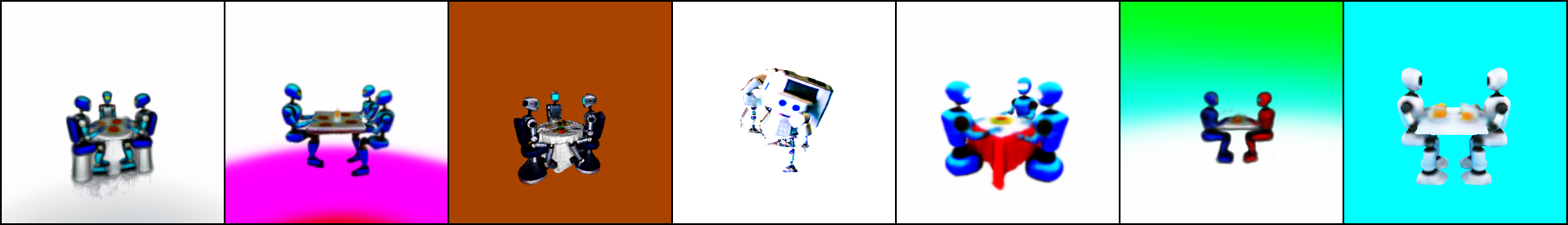}
\centering
\resizebox{1.0\textwidth}{!}{%
\begin{tabularx}{\textwidth}{
  >{\centering\arraybackslash}p{0.12\textwidth}
  >{\centering\arraybackslash}p{0.12\textwidth}
  >{\centering\arraybackslash}p{0.12\textwidth}
  >{\centering\arraybackslash}p{0.12\textwidth}
  >{\centering\arraybackslash}p{0.12\textwidth}
  >{\centering\arraybackslash}p{0.12\textwidth}
  >{\centering\arraybackslash}X
}
\centering DreamFusion-IF \cite{DreamFusion, threestudio} & 
\centering TextMesh-IF \cite{TextMesh, threestudio} & 
\centering Magic3D-IF \cite{Magic3D, threestudio} & 
\centering Fantasia3D \cite{Fantasia3D, threestudio} & 
\centering\arraybackslash AToM Per-prompt (Ours) &
\centering ATT3D-IF$^{\dagger}$ \cite{ATT3D}& 
\centering\arraybackslash \textbf{AToM (Ours)}\\
\end{tabularx}%
}
\caption{\textbf{Visual comparisons} of AToM against the state-of-the-art per-prompt solutions (first four columns), AToM Per-prompt, and our reproduced ATT3D in DF27 dataset. AToM achieves higher quality than ATT3D and a performance comparable to the per-prompt solutions.}
\label{fig:sota}
\end{figure*}

%% file: figures/gallery2.tex
\begin{figure*}[h]
\centering
\includegraphics[trim={0 9cm 0 0},clip,width=1.0\textwidth]{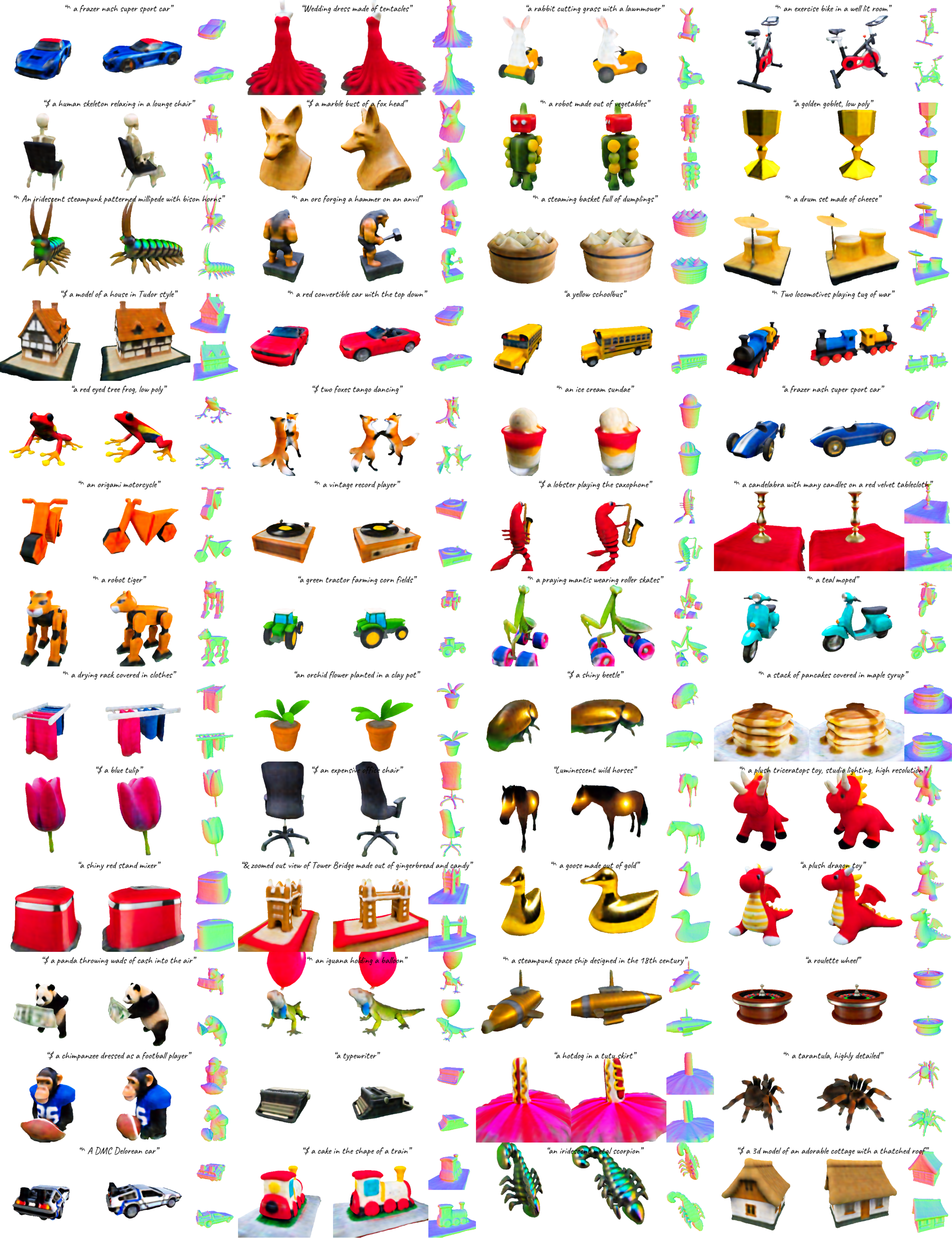}
\vspace{-2em}
\caption{\textbf{More results of AToM} evaluated in DF415.}
\label{fig:gallery2}
\end{figure*}